\documentclass[a4paper]{llncs}

\usepackage{amssymb}
\usepackage{amsmath}
\usepackage{booktabs}
\usepackage{caption}
\usepackage{color}
\usepackage{enumitem}
\usepackage{fancyhdr}
\usepackage{graphicx}
\usepackage{listings}
\usepackage{subfigure}
\usepackage{svn-multi}
\usepackage{tabularx}
\usepackage{times}
\usepackage{url}
\usepackage[table]{xcolor}% http://ctan.org/pkg/xcolor

\graphicspath{{./Figures/}}

\AtBeginDocument{%
  \mathchardef\mathcomma\mathcode`\,
  \mathcode`\,="8000
}
{\catcode`,=\active
  \gdef,{\mathcomma\discretionary{}{}{}}
}
\mathchardef\breakingcomma\mathcode`\,
{\catcode`,=\active
  \gdef,{\breakingcomma\discretionary{}{}{}}
}
\newcommand{\mathlist}[1]{\ensuremath{\mathcode`\,=\string"8000 #1}}

\newcommand{\afst}{\ensuremath{\alpha}}
\newcommand{\alst}{\ensuremath{\omega}}
\newcommand{\bag}[1]{\ensuremath{{\cal B}(#1)}}
\newcommand{\cdf}[1]{\ensuremath{\cnt{#1}{df}}}
\newcommand{\cmax}[1]{\ensuremath{\cnt{#1}{max}}}
\newcommand{\cmin}[1]{\ensuremath{\cnt{#1}{min}}}
\newcommand{\cnt}[2]{\ensuremath{C^{\mathrm{#2}}_{#1}}}
\newcommand{\csum}[1]{\ensuremath{\cnt{#1}{sum}}}
\newcommand{\ebag}[1]{\ensuremath{\left[\mathlist{#1}\right]}}
\newcommand{\eseq}[1]{\ensuremath{\left\langle\mathlist{#1}\right\rangle}}
\newcommand{\eset}[1]{\ensuremath{\left\{\mathlist{#1}\right\}}}
\newcommand{\ext}[2]{\ensuremath{\overline{{#1}_{#2}}}}
\newcommand{\first}[1]{\ensuremath{\uparrow\!\!({#1})}}
\newcommand{\fltr}[3]{{#1}\!\pm\!({#2},{#3})}
\newcommand{\last}[1]{\ensuremath{\downarrow\!\!({#1})}}
\newcommand{\mtch}[2]{{#1}\sqsupseteq{#2}}
\newcommand{\Nat}{\mathrm{I\kern-1.5pt N}}
\newcommand{\occs}[2]{\ensuremath{{#1}\#{#2}}}
\newcommand{\proj}[2]{\ensuremath{{#1}\!\!\upharpoonright\!\!{#2}}}
\newcommand{\raa}[1]{\ensuremath{\rel{#1}{aa}}}
\newcommand{\rab}[1]{\ensuremath{\rel{#1}{ab}}}
\newcommand{\rdf}[1]{\ensuremath{\rel{#1}{df}}}
\newcommand{\rel}[2]{\ensuremath{R^{\mathrm{#2}}_{#1}}}
\newcommand{\replace}[2]{#2}
\newcommand{\req}[1]{\ensuremath{\rel{#1}{eq}}}
\newcommand{\rnt}[1]{\ensuremath{\rel{#1}{nt}}}
\newcommand{\seq}[1]{\ensuremath{{#1}^*}}
\newcommand{\skl}[1]{\ensuremath{{\cal S}(#1)}}
\newcommand{\ua}{\ensuremath{{\cal A}}}
\newcommand{\qntr}[3]{\ensuremath{\mathop{#1}_{#2}{#3}}}
\newcommand{\bnr}[3]{\ensuremath{\left(#1\right) #2 \left(#3\right)}}
\newcommand{\card}[1]{\left|{#1}\right|}

\begin{document}

\mainmatter

\title{Log Skeletons: A Classification Approach to Process Discovery}
\titlerunning{Log Skeletons}
\toctitle{Log Skeletons}

\author{H.M.W. Verbeek \and R.M. de Carvalho}
\authorrunning{H.M.W. Verbeek \and R.M. de Carvalho}
\tocauthor{H.M.W. Verbeek and R.M. de Carvalho}

\institute{
    Department of Mathematics and Computer Science\\
    Eindhoven University of Technology, Eindhoven, The Netherlands\\
    \url{H.M.W.Verbeek@tue.nl}, \url{R.Medeiros.de.Carvalho@tue.nl}\\
}

\maketitle

\begin{abstract}
To test the effectiveness of process discovery algorithms, a Process Discovery Contest (PDC) has been set up.
This PDC uses a classification approach to measure this effectiveness:
The better the discovered model can classify whether or not a new trace conforms to the event log, the better the discovery algorithm is supposed to be.
Unfortunately, even the state-of-the-art fully-automated discovery algorithms score poorly on this classification.
Even the best of these algorithms, the Inductive Miner, \replace{scores}{scored} only $147$ correct classified traces out of $200$ traces on the PDC of 2017.
This paper introduces the \replace{}{rule-based }log skeleton model, which is closely related to \replace{a}{the} Declare constraint model, together with a way to classify traces using this model.
This classification using log skeletons is shown to score better on the PDC of 2017 than state-of-the-art discovery algorithms: $194$ out of $200$.
As a result, one can argue that the fully-automated algorithm to construct \replace{}{(or: \emph{discover}) }a log skeleton from an event log outperforms \replace{}{existing }state-of-the-art fully-automated discovery algorithms.
\end{abstract}

\section{Introduction}\label{sec:intro}

In the area of \emph{process mining}~\cite{pmbook2}, we typically distinguish three main fields: \emph{process discovery}, \emph{process conformance}, and \emph{process enhancement}.
Process discovery deals with discovering a process model from an event log.
Process conformance checks the conformance between a process model and an event log.
Process enhancement enriches a process model using an event log with, for example, resource (who did what?) or timing (how long did it take?) information.

In the field of \emph{process discovery}, many discovery algorithms have been proposed in the past.
Example of such process discovery algorithms include  the Alpha Miner~\cite{alpha}, the ILP Miner~\cite{ilp,hybridilp}, the Heuristics Miner~\cite{heuristics}, the Declare Miner~\cite{declare}, and the Inductive Miner~\cite{im}.
All these algorithms aim to discover some process model from an event log, where the process model discovered depends on the algorithm used.
For example, the Alpha Miner discovers a workflow net~\cite{wfnets}, the ILP Miner a Petri net~\cite{petrinets}, the Heuristics Miner a heuristics net~\cite{heuristics}, \replace{de}{the} Declare Miner a \replace{}{rule-based }constraint model~\cite{declare}, and the Inductive Miner a process tree~\cite{pmbook2}.

To test the effectiveness of the different discovery algorithms, recently, a \emph{Process Discovery Contest} (PDC)~\cite{pdc} has been set up.
The goal of the PDC is to check which of the existing discovery algorithms yield the best process models.
To achieve this goal, the PDC typically contains $10$ different event logs from which a process model needs to be discovered.
To be able to decide which discovery algorithm yields better models, for every event log a set of $20$ traces \replace{are}{is} provided.
For these $20$ traces it is known that $10$ traces are from the same process as the event log, and that $10$ traces are not.
The better the discovered model classifies these $20$ traces into positive and negative traces, the better the discovery algorithm \replace{must}{is supposed to} be.

The typical approach for this classification problem is to use the results from the \emph{process conformance} field:
We simply check which of these $20$ traces conform to the discovered model.
These conforming traces are then classified as positive, while the others are classified as negative.

In this paper, we reverse the roles of the process discovery and the classification:
If having good process discovery algorithm results in having good classifications, then a having good classifications results in having a good discovery algorithm.
As a result, instead of \replace{trying to create}{aiming for} another discovery algorithm that classifies well, we \replace{try to create}{aim for} a \replace{}{good }classification algorithm and assume that this discovers well.

The models that are automatically constructed for the classification algorithm as proposed by this paper are called \emph{log skeletons}, and they are closely related to the Declare constraint models.
In fact, these log skeletons include a number of Declare constraints~\cite{declare}, but they also include some new constraints that are not found in Declare.
The \replace{result}{results} of the PDC of 2017 show\replace{,}{} that with this fully automated classification algorithm we outplay all automated discovery algorithms.
Of the $200$ traces that needed to be classified, the participating automated discovery algorithms (which includes the Inductive Miner) classified at most $153$ traces correctly, whereas our classification algorithm classifies $194$ traces correctly.
As a result of this, the algorithm that constructs the log skeleton model could be considered to be a very good discovery algorithm.

The remainder of this paper is organized as follows.
First, Section~\ref{sec:prelim} introduces the necessary concepts for the section to follow, like \emph{activity logs} etc.
Note that in this paper we restrict event logs to activity logs, as the only information we use from the event is the name of the activity involved.
By taking an event log, and by replacing every event with the involved activity name, we obtain an activity log.
Second, Section~\ref{sec:skelet} defines the log skeleton model.
Third, Section~\ref{sec:vis} shows how the log skeleton models are visualized to the user, which allows the user to inspect the \emph{discovered} model.
Fourth, Section~\ref{sec:class} defines how the log skeleton models of an activity log are used to classify the traces as positive or negative.
Fifth, Section~\ref{sec:impl} shows how the entire approach has been implemented in ProM 6~\cite{prom6}.
Sixth, Section~\ref{sec:contest} shows the promising results of our implemented approach on the PDC of 2017.
Last, Section~\ref{sec:conc} concludes the paper.

\section{Preliminaries}\label{sec:prelim}

Although an event log can conceptually be seen as a set of sequences of events (as every event can be assumed to be unique), an activity log needs to be a \emph{bag} (or multi-set) of sequences of activities.

\begin{definition}[Bags]
If $S$ is a set, then $\bag{S}$ denotes the set of all bags over $S$.
For a $B \in \bag{S}$ and a $s \in S$, $B(s)$ denotes the number of times $s$ occurs in $B$ (often called the cardinality of $s$ in $B$).
Note that the set $S$ can also be considered to be a bag over $S$, namely the bag $B$ such that $B(s) = 1$ if $s \in S$ and $B(s) = 0$ otherwise.
\end{definition}
We use $\ebag{x, y^2, z}$ to denote the bag containing one element $x$, two elements $y$, and one element $z$.
We use $\ebag{}$ to denote the empty bag.

Let $\ua$ be the universe of activities.
The example set of activities $A_1 \in \ua$ contains the activities $a_1, \ldots, a_8$, that is, $A_1=\eset{a_1, \ldots, a_8}$.

\begin{definition}[Activity trace]
An activity trace $\sigma$ over a set of activities $A \subseteq \ua$ is a sequence of activities, that is, $\sigma \in \seq{A}$.
\end{definition}
We use $\eseq{x, y, z, y}$ to denote the \replace{bag}{sequence} containing first an element $x$, second an element $y$, third an one element $z$, and fourth and last another element $y$.
An example trace $\sigma_1$ over $A_1$ is $\sigma_1 = \eseq{a_1, a_2, a_4, a_5, a_6, a_2, a_4, a_5, a_6, a_4, a_2, a_5, a_7}$.
We use $\eseq{}$ to denote the empty \replace{bag}{sequence}.

A sequence can be projected on a subset of its elements in the usual way: $\proj{\sigma_1}{\eset{a_1, a_7, a_8}} = \eseq{a_1, a_7}$.
We use $\first{\sigma}$ and $\last{\sigma}$ to denote the first and the last activity in the (non-empty) trace $\sigma$.
As examples, $\first{\sigma_1} = a_1$ and $\last{\sigma_1} = a_7$.
Furthermore, we use $\card{\sigma}$ to count the elements in the trace $\sigma$ and we use $\occs{\sigma}{\sigma'}$ to count how often the non-empty trace $\sigma'$ occurs as subtrace in the trace $\sigma$.
As examples, $\card{\sigma_1} =13$, $\occs{\sigma_1}{\eseq{a_2,a_4,a_5}} = 2$, $\occs{\eseq{}}{\eseq{a_2,a_4,a_5}} = 0$, and $\occs{\sigma_1}{\eseq{a_6,a_7}} = 0$.

\begin{definition}[Activity log]
An activity log $L$ over a set of activities $A \subseteq \ua$ is a bag of activity traces over $A$, that is, $L \in \bag{\seq{A}}$.
\end{definition}
\begin{table*}[tb]
\centering
\caption{Activity log $L_1$ in tabular form.}
\label{tab:prelim:L1}
\begin{tabular}{|rl|c|}
\toprule
\multicolumn{2}{|l|}{\textbf{Trace}} & \textbf{Frequency} \\ \midrule
$\sigma_1=$ & $\eseq{a_1,a_2,a_4,a_5,a_6,a_2,a_4,a_5,a_6,a_4,a_2,a_5,a_7}$ & 1 \\
$\sigma_2=$ & $\eseq{a_1,a_2,a_4,a_5,a_6,a_3,a_4,a_5,a_6,a_4,a_3,a_5,a_6,a_2,a_4,a_5,a_7}$ & 1 \\
$\sigma_3=$ & $\eseq{a_1,a_2,a_4,a_5,a_6,a_3,a_4,a_5,a_7}$ & 1 \\
$\sigma_4=$ & $\eseq{a_1,a_2,a_4,a_5,a_6,a_3,a_4,a_5,a_8}$ & 2 \\
$\sigma_5=$ & $\eseq{a_1,a_2,a_4,a_5,a_6,a_4,a_3,a_5,a_7}$  & 1 \\
$\sigma_6=$ & $\eseq{a_1,a_2,a_4,a_5,a_8}$ & 4 \\
$\sigma_7=$ & $\eseq{a_1,a_3,a_4,a_5,a_6,a_4,a_3,a_5,a_7}$ & 1 \\
$\sigma_8=$ & $\eseq{a_1,a_3,a_4,a_5,a_6,a_4,a_3,a_5,a_8}$ & 1 \\
$\sigma_9=$ & $\eseq{a_1,a_3,a_4,a_5,a_8}$ & 1 \\
$\sigma_{10}=$ & $\eseq{a_1,a_4,a_2,a_5,a_6,a_4,a_2,a_5,a_6,a_3,a_4,a_5,a_6,a_2,a_4,a_5,a_8}$ & 1 \\
$\sigma_{11}=$ & $\eseq{a_1,a_4,a_2,a_5,a_7}$ & 3 \\
$\sigma_{12}=$ & $\eseq{a_1,a_4,a_2,a_5,a_8}$ & 1 \\
$\sigma_{13}=$ & $\eseq{a_1,a_4,a_3,a_5,a_7}$ & 1 \\
$\sigma_{14}=$ & $\eseq{a_1,a_4,a_3,a_5,a_8}$ & 1 \\\bottomrule
\end{tabular}
\end{table*}
Table~\ref{tab:prelim:L1} shows an example activity log over $A_1$, which contains 20 traces and 14 different traces.

An activity log $L$ over a set of activities $A \subseteq \ua$ can be projected on a subset of activities, which projects every trace in the log to that subset of activities.
As an example, $\proj{L_1}{\eset{a_1, a_7, a_8}} = \ebag{\eseq{a_1, a_7}^9, \eseq{a_1, a_8}^{11}}$.

\section{Log Skeleton}\label{sec:skelet}

For the log skeleton, we extend every activity trace with an artificial start activity $\afst$ and an artificial end activity $\alst$, as we believe making the start and end of an activity trace explicit gives a better picture of the activity log in the end.
For this reason, we introduce the concepts of extended traces and extended logs.

\begin{definition}[Artificial start and end activity]
The activity $\afst \in \ua$ is an artificial activity that denotes the start of a trace.
Likewise, the activity $\alst \in \ua$ is an artificial activity that denotes the end of a trace.
\end{definition}

\begin{definition}[Extended set of activities]
Let $A \subseteq \ua$ be a set of activities such that $\afst,\alst \in \ua \setminus A$.
Then $\ext{A}{} = A \cup \eset{\afst,\alst}$ is the extended set of activities of $A$.
\end{definition}
As an example, $\ext{A}{1} = \eset{\afst,a_1,a_2,a_3,a_4,a_5,a_6,a_7,a_8,\alst}$.

\begin{definition}[Extended trace]
Let $\sigma = \eseq{a,\ldots,a'}$ be an activity trace over a set of activities $A$ such that $\afst,\alst \in \ua \setminus A$.
Then $\ext{\sigma}{} = \eseq{\afst,a,\ldots,a',\alst}$ is the \emph{extended} trace of $\sigma$.
\end{definition}
As an example, $\ext{\sigma}{1} = \eseq{\afst,a_1,a_2,a_4,a_5,a_6,a_2,a_4,a_5,a_6,a_4,a_2,a_5,a_7,\alst}$.

\begin{definition}[Extended log]
Let $L$ be an activity log over some set of activities $A \in \ua$ such that $\afst,\alst \in \ua \setminus A$.
Then $\ext{L}{} = \ebag{\ext{\sigma}{}\left|\sigma \in L\right.}$ is the \emph{extended} log of $L$.
\end{definition}

Having defined the extended log with extended traces, we can now define the main model in this paper: log skeletons.

\begin{definition}[Log skeleton]
Let $L$ be an activity log over some set of activities $A \in \ua$.
The log skeleton of $L$ is denoted $\skl{L}$ and is defined as $(\req{L},\raa{L},\rab{L},\rnt{L},\rdf{L},\cdf{L},\csum{L},\cmin{L},\cmax{L})$, where:
\begin{itemize}
\item $\req{L} \subseteq (\ext{A}{} \times \ext{A}{})$ is an equivalence relation such that
\[{\bnr{(a,a') \in \req{L}}{\Leftrightarrow}{\qntr{\forall}{\sigma \in \ext{L}{}}{\card{\proj{\sigma}{\eset{a}}} = \card{\proj{\sigma}{\eset{a'}}}}}},\]
that is, two activities are related by $\req{L}$  if and only if they occur equally often in every trace.
This equivalence relation has \replace{not}{no} direct counterpart in Declare, but it is straightforward that this relation implies the \emph{co-existence} constraint.
\item $\raa{L} \subseteq (\ext{A}{} \times \ext{A}{})$ is a transitive and non-reflexive \emph{always-after} relation such that
\[{\bnr{(a,a') \in \raa{L}}{\Leftrightarrow}{\qntr{\forall}{\sigma \in \ext{L}{}}{\left(\proj{\sigma}{\eset{a}}=\eseq{}\right) \vee \left(\last{\proj{\sigma}{\eset{a, a'}}} = a'\right)}}},\]
that is, two activities are related by $\raa{L}$ if and only if after any occurrence of the first activity the second activity always occurs.
The \emph{always-after} relation corresponds to the \emph{succession} constraint in Declare.
\item $\rab{L} \subseteq (\ext{A}{} \times \ext{A}{})$ is a transitive and non-reflexive \emph{always-before} relation such that
\[{\bnr{(a,a') \in \rab{L}}{\Leftrightarrow}{\qntr{\forall}{\sigma \in \ext{L}{}}{\left(\proj{\sigma}{\eset{a}}=\eseq{}\right) \vee \left(\first{\proj{\sigma}{\eset{a, a'}}} = a'\right)}}},\]
that is, two activities are related by $\rab{L}$ if and only if before any occurrence of the first activity the second activity always occurs.
The \emph{always-before} relation corresponds to the \emph{precedence} constraint in Declare.
\item $\rnt{L} \subseteq (\ext{A}{} \times \ext{A}{})$ is a symmetric \emph{never-together} relation such that
\[{\bnr{(a,a') \in \rnt{L}}{\Leftrightarrow}{\qntr{\forall}{\sigma \in \ext{L}{}}{(\proj{\sigma}{\eset{a}}=\eseq{}) \vee (\proj{\sigma}{\eset{a'}}=\eseq{})}}},\]
that is, two activities are related by $\rnt{L}$ if and only if they do not occur together in any trace.
The \emph{never-together} relation corresponds to the \emph{non co-existence} constraint in Declare.
\item $\rdf{L} \subseteq (\ext{A}{} \times \ext{A}{})$ is a \emph{directly-follows} relation such that
\[{\bnr{(a,a') \in \rdf{L}}{\Leftrightarrow}{\qntr{\exists}{\sigma \in \ext{L}{}}{\occs{\sigma}{\eseq{a,a'} > 0}}}},\]
that is, two activities are related by $\rdf{L}$ if and only if an occurrence the first activity can directly be followed by an occurrence of the second.
The \emph{directly-follows} relation has no counterpart in Declare.
\item $\cdf{L} \in (\ext{A}{} \times \ext{A}{}) \rightarrow \Nat$ is a \emph{directly-follows} counter such that
\[{\cdf{L}(a,a') = \qntr{\sum}{\sigma \in \ext{L}{}}{\occs{\sigma}{\eseq{a,a'}}}},\]
that is, $\cdf{L}$ returns for every pair of activities how often an occurrence of the first activity is directly followed by an occurrence of the second in the entire log.
\item $\csum{L} \in \ext{A}{} \rightarrow \Nat$ is a \emph{sum} counter such that
\[{\csum{L}(a) = \qntr{\sum}{\sigma \in \ext{L}{}}{|\proj{\sigma}{\eset{a}}|}},\]
that is, $\csum{L}$ returns for every activity how often this activity occurs in the entire log.
\item $\cmin{L},\cmax{L} \in \ext{A}{} \rightarrow \Nat$ are \emph{min} and \emph{max} counters such that
\[{\bnr{\cmin{L}(a) = \qntr{\min}{\sigma \in \ext{L}{}}{\card{\proj{\sigma}{\eset{a}}}}}{\wedge}{\cmax{L}(a) = \qntr{\max}{\sigma \in \ext{L}{}}{\card{\proj{\sigma}{\eset{a}}}}}},\]
that is, $\cmin{L}$ ($\cmax{L}$) returns for every activity the minimal (maximal) number of occurrences of this activity in any trace.
Together, $\cmin{L}$ and $\cmax{L}$ are related to the \emph{existence}, \emph{absence}, and \emph{exactly} constraints in Declare.
\end{itemize}
\end{definition}
\replace{As examples:
\begin{itemize}
\item $(a_4,a_5) \in \req{L_1}$ and $(\afst, a_1) \in \req{L_1}$, that is, in any extended trace $a_4$ and $a_5$ occur equally often, and $a_1$ occurs exactly once (as, by definition, $\afst$ occurs exactly once).
\item $(a_2,a_5) \in \raa{L_1}$, $(a_1,a_4) \in \raa{L_1}$, and $(a_1,\alst) \in \raa{L_1}$, that is, in any extended trace, after any occurrence of $a_2$ there is always an occurrence of $a_5$, etc.
\item $(a_6,a_5) \in \rab{L_1}$, $(a_4,a_1) \in \rab{L_1}$, and $(a_1,\afst) \in \rab{L_1}$, that is, in any extended trace, before any occurrence of $a_6$ there is always an occurrence of $a_5$, etc.
\item $(a_7,a_8) \in \rnt{L_1}$, that is, $a_7$ and $a_8$ never occur together in a single extended trace.
\item $(a_1,a_2) \in \rdf{L_1}$, $(a_2,a_4) \in \rdf{L_1}$, and $(a_4,a_2) \in \rdf{L_1}$, that is, in some extended trace (like $\sigma_1$) $a_1$ is directly followed by $a_2$, etc.
\item $\cdf{L_1}((a_1,a_2)) = 10$, $\cdf{L_1}((a_2,a_1)) = 0$, $\cdf{L_1}((a_2,a_4)) = 13$, and $\cdf{L_1}((a_4,a_2)) = 7$, that is, in the extended log $a_1$ is 10 times directly followed by $a_2$, etc.
\item $\csum{L_1}(\afst) = \csum{L_1}(\alst) = \csum{L_1}(a_1) = 20$ and $\csum{L_1}(a_4) = \csum{L_1}(a_5) = 34$, that is, in the extended log $\afst$, $\alst$, and $a_1$ occur exactly 20 times, whereas $a_4$ and $a_5$ occur exactly 34 times.
\item $\cmin{L_1}(\afst) = \cmin{L_1}(\alst) = \cmin{L_1}(a_1) = 1$ and $\cmin{L_1}(a_4) = \cmin{L_1}(a_5) = 1$, that is, in any extended trace $\afst$, $\alst$, and $a_1$ occur minimally 1 time, and $a_4$ and $a_5$ also occur minimally 1 time.
\item $\cmax{L_1}(\afst) = \cmax{L_1}(\alst) = \cmax{L_1}(a_1) = 1$ and $\cmax{L_1}(a_4) = \cmax{L_1}(a_5) = 4$, that is, in the extended log $\afst$, $\alst$, and $a_1$ occur maximally 1 time, whereas $a_4$ and $a_5$ occur maximally 4 times.
\end{itemize}}{}

\replace{By definition, it is clear that the following properties hold for a log skeleton $S = \skl{L}$:
\begin{itemize}
\item $\forall_{(a,a') \in \req{L}{}}[\csum{L}(a) = \csum{L}(a')]$, that is, equivalent activities occur equally often in the extended log.
\item $\forall_{(a,a') \in \req{L}{}}[\cmin{L}(a) = \cmin{L}(a') \wedge \cmax{L}(a) = \cmax{L}(a')]$, that is, equivalent activities occur equally often in any extended trace.
\item $\csum{L}(\afst) = \csum{L}(\alst) = |L|$, that is, the artificial activities $\afst$ and $\alst$ occur as many times as the extended log contains traces.
\item $\cmin{L}(\afst) = \cmax{L}(\afst) = \cmin{L}(\alst) = \cmax{L}(\alst) = 1$, that is, the  artificial activities $\afst$ and $\alst$ occur exactly once in any extended trace.
\item $\forall_{(a,a') \in \rnt{L}{}}[(a,a') \not\in \req{L}{} \cup \raa{L}{} \cup \rab{L}{} \cup \rdf{L}{}]$, that is, if two activities never occur together in any trace, they cannot be equivalent, they can not always be after or before, nor can they directly follow each other.
\item $(|L| = 0) \Rightarrow (\req{L} = \raa{L} = \rab{L} = \rnt{L} = (\ext{A}{} \times \ext{A}{}) \wedge \rdf{L} = \emptyset)$, that is, if the log is empty, then all activities are equivalent, every two activities are always-after, always-before and never-together, and no two activities are directly-follows.
\end{itemize}}{}

\section{Visualization}\label{sec:vis}

We will visually represent a log skeleton as graph, where the extended set of activities are the nodes of the graph, and the relations are the edges of the graph.

\begin{figure}[tb]
\centering
\includegraphics[width=0.886\textwidth,trim={0px 80px 0px 10px},clip]{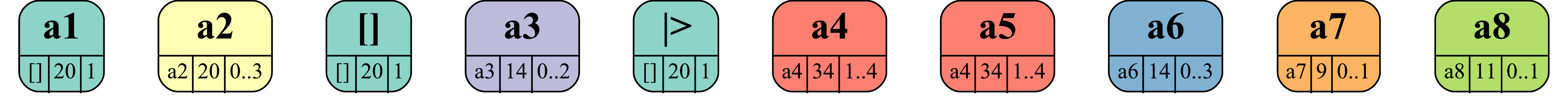}
\caption{The visual representation of the activities for activity log $L_1$, which also shows the equivalence relation $\req{L_1}$.}
\label{fig:skltn:req}
\end{figure}
Figure~\ref{fig:skltn:req} shows the visual representation of the extended activities for the log $L_1$, where $|\!\!>$ denotes $\afst$ and $[]$ denotes $\alst$.
The top of the node $a$ contains the activity name, that is, $a$.
The bottom of the node $a$ contains, from left to right:
\begin{itemize}
\item $a'$, where $\bnr{(a,a') \in \req{L}{}}{\wedge}{\qntr{\forall}{a'' \in \ext{A}{}}{\bnr{(a,a'') \in \req{L}{}}{\Rightarrow}{a' \leq a''}}}$, that is, $a'$ is the smallest (in a lexicographical way) equivalent activity.
\item $\csum{L}{}(a)$, that is, the number of times $a$ has occurred in the extended log.
\item $\cmin{L}{}(a)..\cmax{L}{}(a)$, that is, the interval with the minimal and maximal numbers of times the activity has occurred in any extended trace. If $\cmin{L}{}(a)=\cmax{L}{}(a)$, we simplify this interval to $\cmin{L}{}(a)$.
\end{itemize}
The color gradient of the node also indicates the equivalence class: Different color gradients indicate different equivalence classes.
If we run out of color gradients, or if color gradients are hard to distinguish, the smallest equivalent activity can still be used to decide whether two activities are equivalent: Two activities are equivalent if and only if they have the same smallest equivalent activity.

Both the always-after relation \raa{L} and the always-before relation $\rab{L}$ are visualized after a transitive reduction on these relations, as this possibly removes a lot of redundant edges in the graph.

\begin{figure}[tb]
\centering
\subfigure[Always-after relation $\raa{L_1}$]{%
\label{fig:skltn:raa}%
\includegraphics[width=0.498\textwidth,trim={0px 80px 0px 10px},clip]{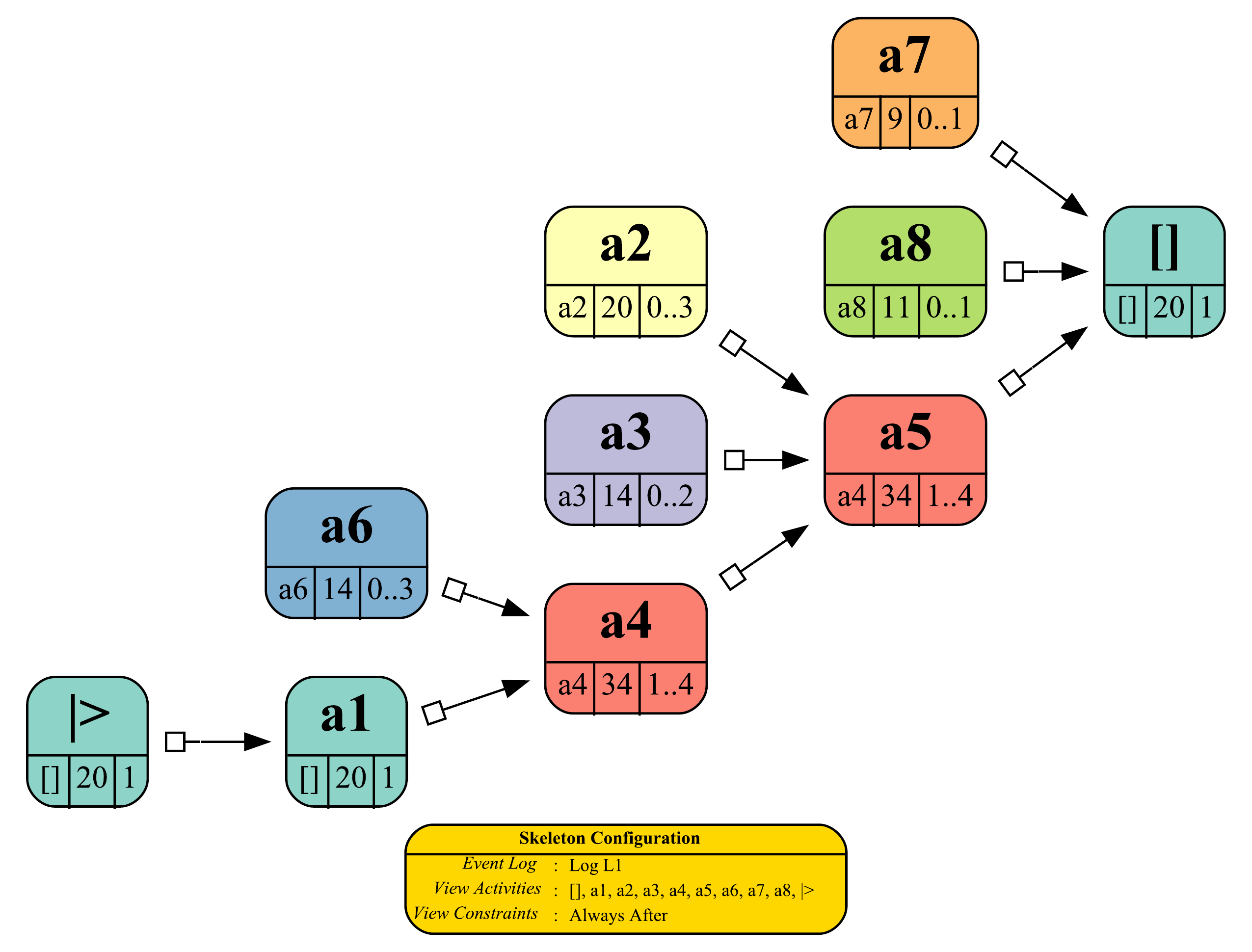}}%
\subfigure[Always-before relation $\rab{L_1}$]{%
\label{fig:skltn:rab}%
\includegraphics[width=0.498\textwidth,trim={0px 80px 0px 10px},clip]{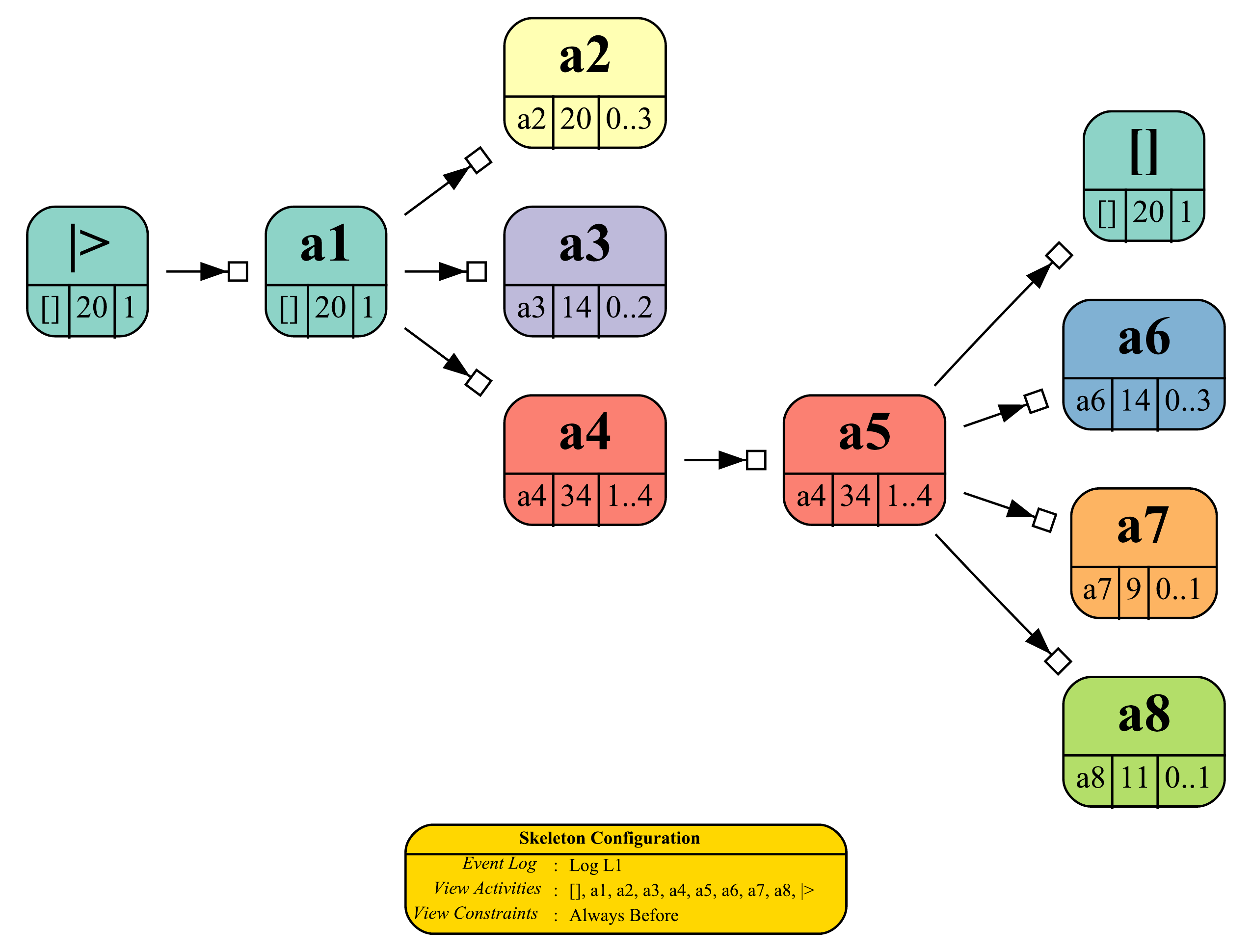}}%
\caption{The visual representations of both always relations for activity log $L_1$.}
\label{fig:skltn:raarab}%
\end{figure}
\replace{\begin{figure}[tb]
\centering
\begin{minipage}{0.498\textwidth}
\includegraphics[width=\textwidth,trim={0px 80px 0px 10px},clip]{raaL1}
\caption{The visual representation of the always-after relation $\raa{L_1}$ for activity log $L_1$.}
\label{fig:skltn:raa}
\end{minipage}%
\begin{minipage}{0.498\textwidth}
\includegraphics[width=\textwidth,trim={0px 80px 0px 10px},clip]{rabL1}
\caption{The visual representation of always-before relation $\rab{L_1}$ for activity log $L_1$.}
\label{fig:skltn:rab}
\end{minipage}
\end{figure}}{}
Figure~\ref{fig:skltn:raarab} shows the visual representation of both always relations for the log $L_1$.
As examples, activity $a_4$ is always after activity $a_1$, activity $a_5$ is always after activity $a_4$, and, as a result, $a_5$ is always after $a_1$; and activity $a_1$ is always before activity $a_4$, activity $a_4$ is always before activity $a_5$, and, as a result, $a_1$ is always before $a_5$.

Note that in both \replace{representation}{representations} the open box on the edge indicates the point of view for the ``always'' part whereas the direction of the arrow indicates whether it is ``after'' (open box at tail) of ``before'' (open box at head).
As a result, we can combine both \replace{representation}{representations} into a single representation.
\begin{figure}[tb]
\centering
\includegraphics[width=0.610\textwidth,trim={0px 80px 0px 10px},clip]{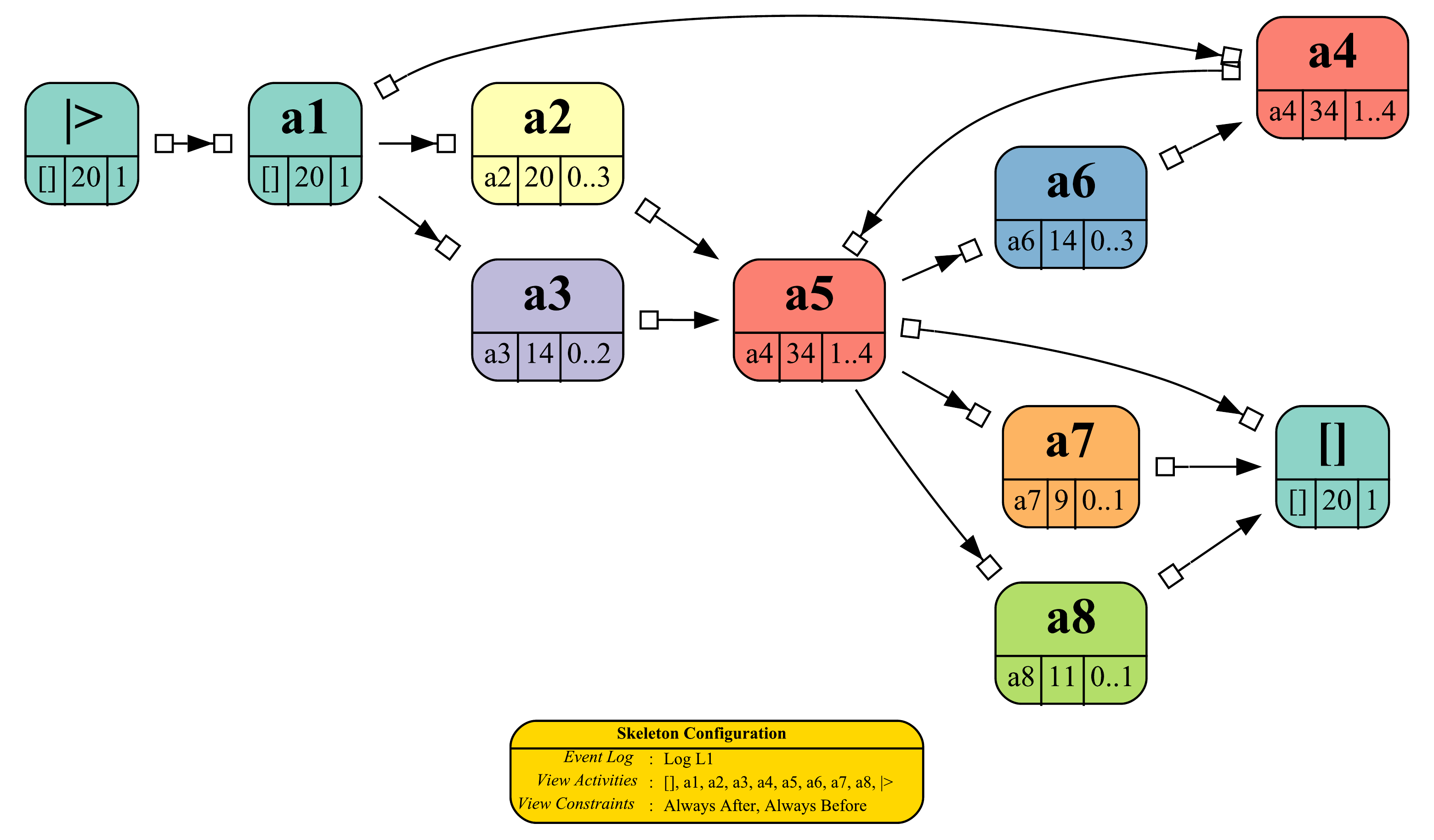}
\caption{The combined visual representation of both always relations for activity log $L_1$.}
\label{fig:skltn:raab}
\end{figure}
Figure~\ref{fig:skltn:raab} shows the combined visual representation of both always relations for the log $L_1$.
In this visualization, the single arc from $a_4$ to $a_5$ captures both the always-after relation (the open box at $a_4$) and the always-before relation (the open box at $a_5$).

\begin{figure}[tb]
\centering
\includegraphics[width=0.797\textwidth,trim={0px 80px 0px 10px},clip]{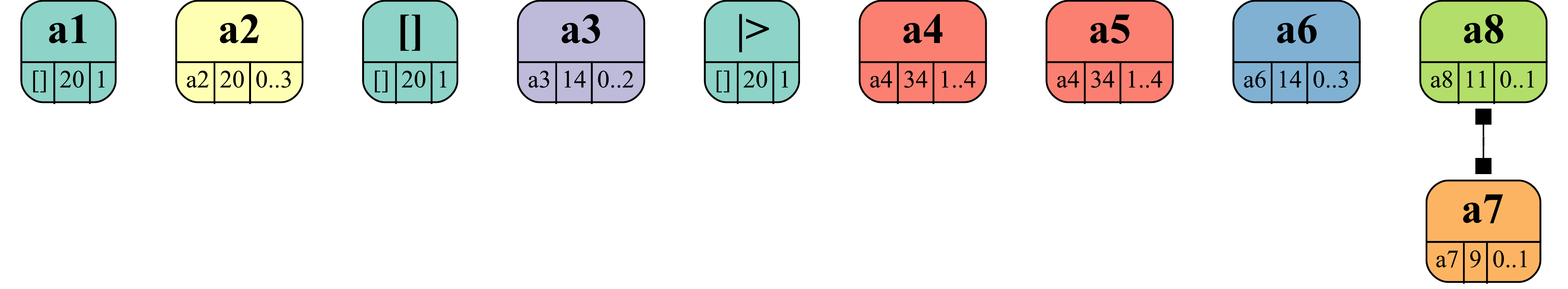}
\caption{The visual representation of the never-together relation $\rnt{L_1}$ for activity log $L_1$.}
\label{fig:skltn:rnt}
\end{figure}
Figure~\ref{fig:skltn:rnt} shows the visual representation of the never-together relation for the log $L_1$.
Apparently, only activities $a_7$ and $a_8$ are related by this.

\begin{figure}[tb]
\centering
\includegraphics[width=0.810\textwidth,trim={0px 80px 0px 10px},clip]{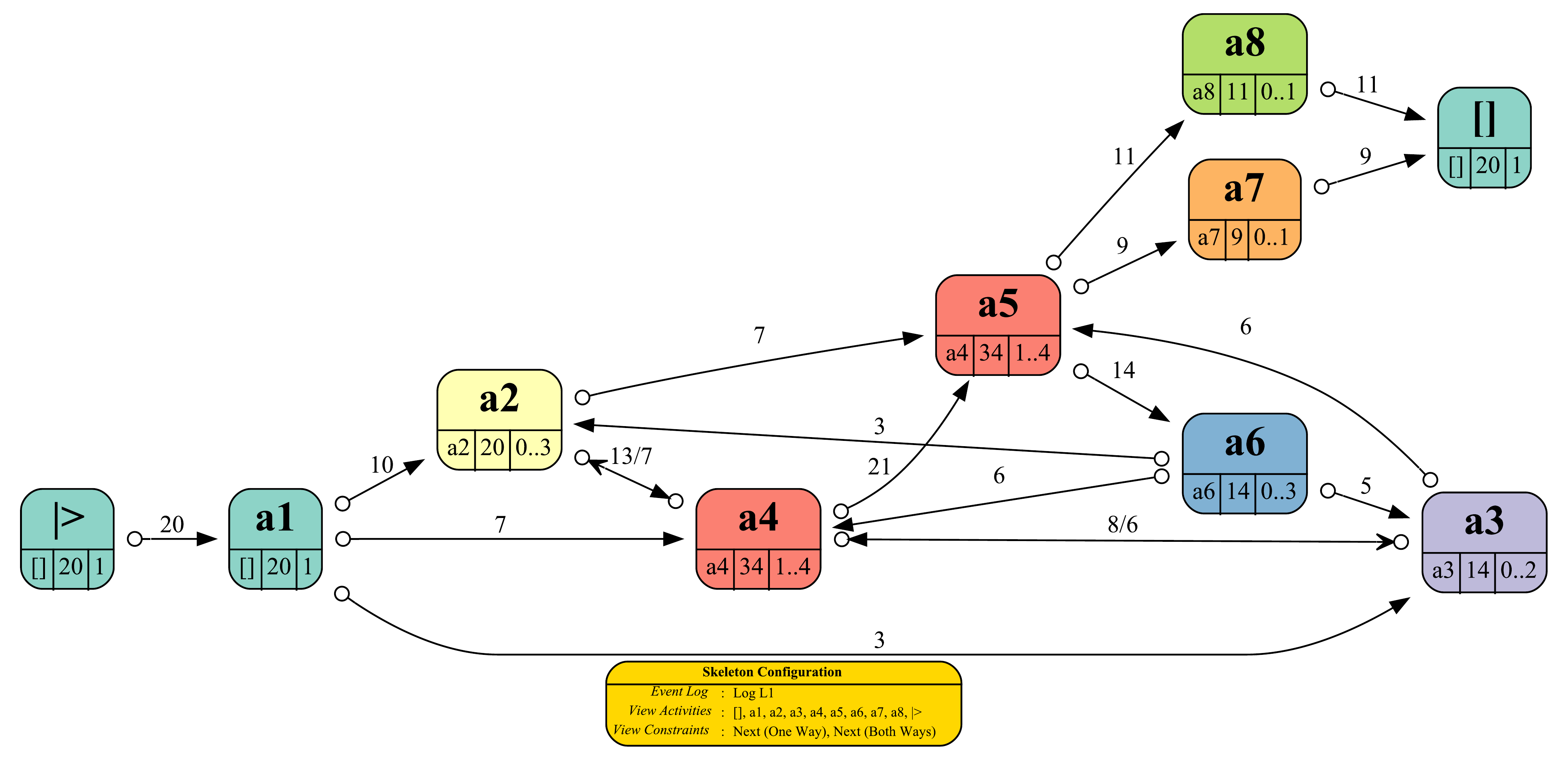}
\caption{The visual representation of the directly-follows relation $\rdf{L_1}$ for activity log $L_1$.}
\label{fig:skltn:rdf}
\end{figure}
Figure~\ref{fig:skltn:rdf} shows the visual representation of the directly follows relation $\rdf{L_1}$ for the log $L_1$, which includes the number of how often in the extended log one activity was directly followed by another.
As examples, activity $a_1$ is 10 times directly followed by activity $a_2$ but never the other way around, while $a_2$ is 13 times directly followed by activity $a_4$ and 7 times the other way around.
Note that for the latter, we used different arc heads to indicate which number belongs to which direction: The first number corresponds to the triangular head while the second number corresponds to the vee-shaped head.

We can combine all these different representations into a single representation.
\begin{figure}[tb]
\centering
\replace{\includegraphics[width=0.786\textwidth,trim={0px 80px 0px 0px},clip]{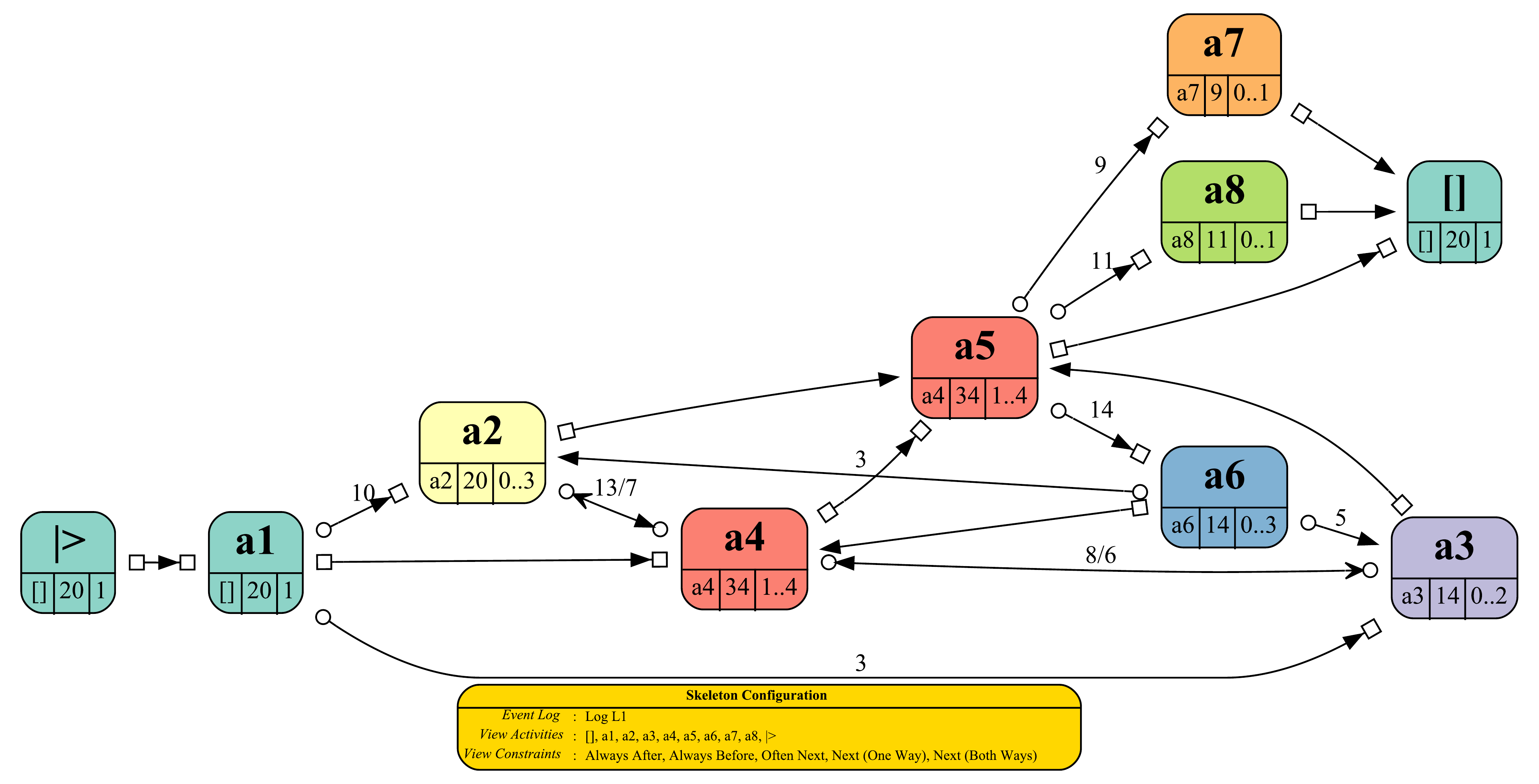}}{\includegraphics[width=0.881\textwidth,trim={0px 80px 0px 10px},clip]{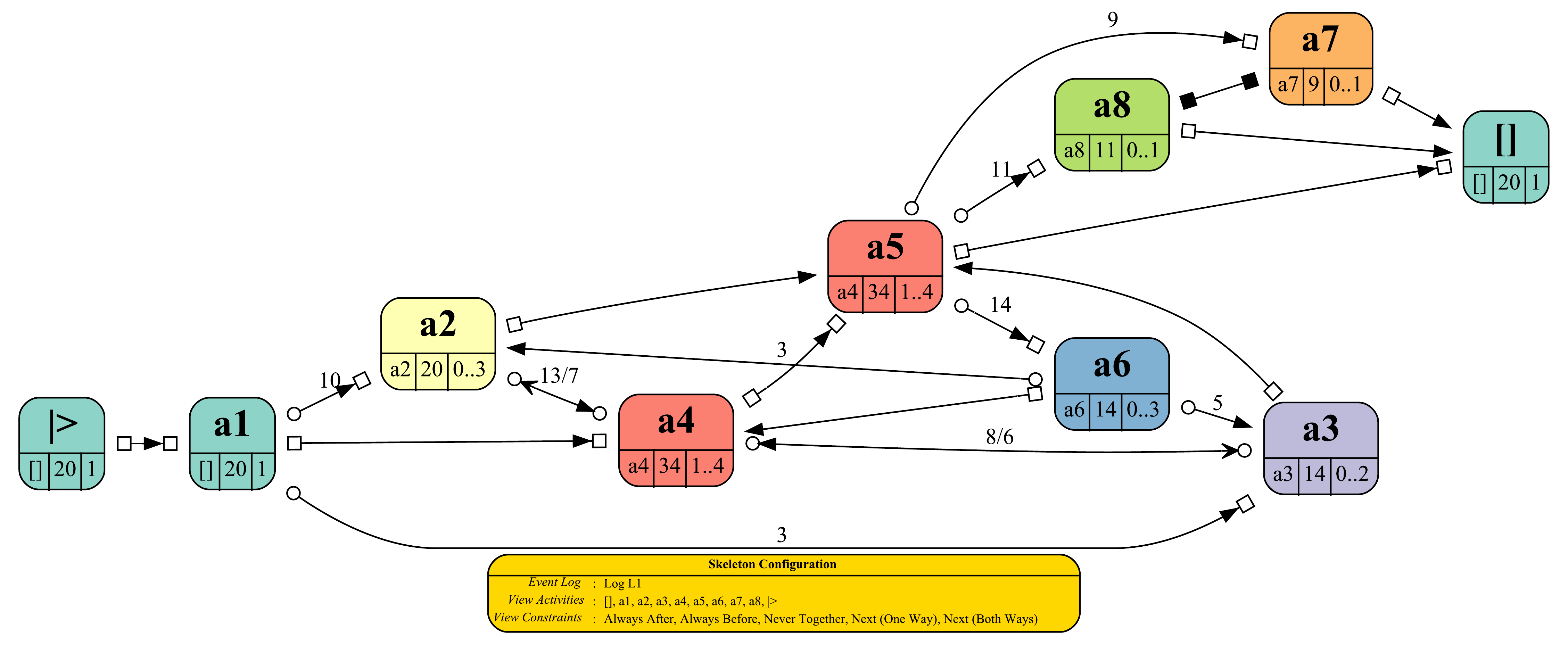}}
\caption{The combined visual representation of the entire log skeleton $\skl{L_1}$ for activity log $L_1$.}
\label{fig:skltn:skl}
\end{figure}
Figure~\ref{fig:skltn:skl} shows the combined visual representation of the entire skeleton $\skl{L_1}$ for the log $L_1$.
In this representation, an always relation may conflict with a directly follows relation from one activity to another.
As an example, after $a_1$ there is always $a_4$, and $a_1$ is directly followed 7 times by $a_4$.
In case of such a conflict the always relation has higher priority, and will cause the directly-follows relation not to be shown.

The default log skeleton for an activity log will, however, not be the log skeleton that shows all relations, as we believe this adds too much clutter (too much edges) in the graph.
Instead, the default log skeleton will contain only both always relations, as these typically provide the most useful information.

\section{Classification}\label{sec:class}

As mentioned in the Introduction, log skeletons are primarily used to \emph{classify} whether some activity trace conforms to an activity log.
An activity trace conforms to an activity log if the trace could have been generated by the same process that generated the log.
Consider, for example, the activity log $L_1$ and the activity trace $\eseq{a_1,a_4,a_5,a_7}$.
Could this trace have been in $L_1$ as well?
To check this, we first define when one log (like $L_1$) subsumes another log (like the log containing only \replace{the}{one} trace \replace{$\sigma$}{$\eseq{a_1,a_4,a_5,a_7}$}).

 \begin{definition}[Log subsumes log]\label{def:sbsmsl}
Let $L$ and $L'$ be two activity logs over some set of activities $A \in \ua$ such that $\afst,\alst \in \ua \setminus A$.
The log $L$ \emph{subsumes} the log $L'$, denoted $\mtch{L}{L'}$,  if and only if the following conditions hold:
\begin{itemize}
\item If two activities are equivalent, always-after, or always-before in the subsuming log, they are also equivalent, always-after, or always-before in the subsumed log:
\[\qntr{\forall}{R \in \eset{\req{},\raa{},\rab{}}}{\left(\qntr{\forall}{a,a' \in \ext{A}{}}{\bnr{(a,a') \in R_{\ext{L}{}}}{\Rightarrow}{(a,a') \in R_{\ext{L'}{}}}}\right)}\]
\item If one activity directly-follows another activity in the subsumed log, it is also directly-followed in the subsuming log:
\[\qntr{\forall}{a,a' \in \ext{A}{}}{\bnr{(a,a') \in \rdf{\ext{L}{}}}{\Leftarrow}{(a,a') \in \rdf{\ext{L'}{}}}}\]
\end{itemize}
\end{definition}
Note that for the last property we assume the so-called $\alpha$-completeness of the log, and that for this property the implication works the other way around.

To be able to effectively check choice constructs, we \replace{will }{}include filtered logs into the subsumption relation.
To explain this in some detail, consider the log $L_1$, and in particular the activities $a_7$ and $a_8$.
It seems obvious that the model contains a mandatory choice between $a_7$ and $a_8$, as the following properties hold:
\begin{itemize}
\item $(a_7,a_8) \in \rnt{L_1}$, that is, $a_7$ and $a_8$ never occur together.
\item $\csum{L_1}(a_7) + \csum{L_1}(a_8) = \card{L_1}$, that is, together $a_7$ and $a_8$ occur as many times as there are traces in the log.
\end{itemize}
From this, we can conclude that in every trace either $a_7$ occurs or $a_8$.
Although for this mandatory choice this conclusion seems rather straightforward, for other mandatory choices this is less straightforward.
For this, consider $a_2$ and $a_3$.
Of course, for these activities, we could argue that $\csum{L_1}(a_2) + \csum{L_1}(a_3) = \csum{L_1}(a_5)$, but why $a_5$ (and why not, say, $a_4$)?

To avoid this, we use a simple filtering scheme, and include this into our subsumption relation.
For example, consider what would happen if we would remove all traces from $L_1$ that contain $a_7$.
The remaining traces would then all contain $a_8$!
As a result, in the filtered log, $a_8$ would become a mandatory activity, and be equivalent to $\afst$ and $\alst$.
Something similar holds for $a_2$ and $a_3$.
If we would remove all traces from $L_1$ that contain $a_2$, then in every remaining trace $a_3$ would happen equally often as $a_5$ (or $a_4$).
The key here is that by filtering, some relations between remaining activities appear that were not there in the entire log.
Instead of having to decide how the equality should look like for some mandatory choice, we can simply remove any set of activities.
At some point in time, we will have removed all alternatives in a mandatory choice, leaving only the remaining choice as a mandatory activity.

 \begin{definition}[Filtered log]
 Let $L$ be an activity log over some set of activities $A \in \ua$ such that $\afst,\alst \in \ua \setminus A$, and let $A^{\mathrm{req}},A^{\mathrm{fbd}} \subseteq A$ such that $A^{\mathrm{req}} \cap A^{\mathrm{fbd}} = \emptyset$.
 The filtered log on both set of activities, denoted $\fltr{L}{A^{\mathrm{req}}}{A^{\mathrm{fbd}}}$, is defined as
 \[\ebag{\sigma\left|\left(\sigma \in L\right) \wedge \left(\qntr{\forall}{a \in A^{\mathrm{req}}}{\proj{\sigma}{\eset{a}} \not= \eseq{}}\right) \wedge \left(\qntr{\forall}{a \in A^{\mathrm{fbd}}}{\proj{\sigma}{\eset{a}} = \eseq{}}\right)\right.},\]
that is, $\fltr{L}{A^{\mathrm{req}}}{A^{\mathrm{fbd}}}$ contains all traces that contain every activity from $A^{\mathrm{req}}$ (the \emph{required} set of activities) and no activity from $A^{\mathrm{fbd}}$ (the \emph{forbidden} set of activities).
\end{definition}

\begin{definition}[Log subsumes trace]\label{def:sbsms}
Let $L$ be an activity log over some set of activities $A \in \ua$ such that $\afst,\alst \in \ua \setminus A$, and let $\sigma$ be an activity trace over the same set of activities.
Log $L$ \emph{subsumes} trace $\sigma$, denoted $\mtch{L}{\sigma}$, if and only if
\[\qntr{\forall}{A^{\mathrm{req}} \subseteq A}{\left(\qntr{\forall}{A^{\mathrm{fbd}} \subseteq (A \setminus A^{\mathrm{req}})}{\mtch{\fltr{L}{A^{\mathrm{req}}}{A^{\mathrm{fbd}}}}{\fltr{\ebag{\sigma}}{A^{\mathrm{req}}}{A^{\mathrm{fbd}}}}}\right)},\]
that is, if and only if for all valid subsets of \emph{required} and \emph{forbidden} activities the projected subsuming log subsumes the projected subsumed log (which contains only one trace $\sigma$).
\end{definition}
As an example, consider the question asked earlier: Is the trace $\eseq{a_1,a_4,a_5,a_7}$ subsumed by the log $L_1$?
It can be checked that the log $\fltr{L_1}{\eset{}}{\eset{}}$ subsumes the log $\fltr{\ebag{\sigma}}{\eset{}}{\eset{}}$.
Hence, without filtering, we would not be able to say that $\sigma$ is not subsumed by $L_1$.
However, if we check whether the log  $\fltr{L_1}{\eset{}}{\eset{a_2}}$ subsumes the log $\fltr{\ebag{\sigma}}{\eset{}}{\eset{a_2}}$, that is, with $a_2$ as the only forbidden activity, we would be able to say so.
\begin{figure}[tb]
\centering
\includegraphics[width=0.571\textwidth,trim={0px 95px 0px 10px},clip]{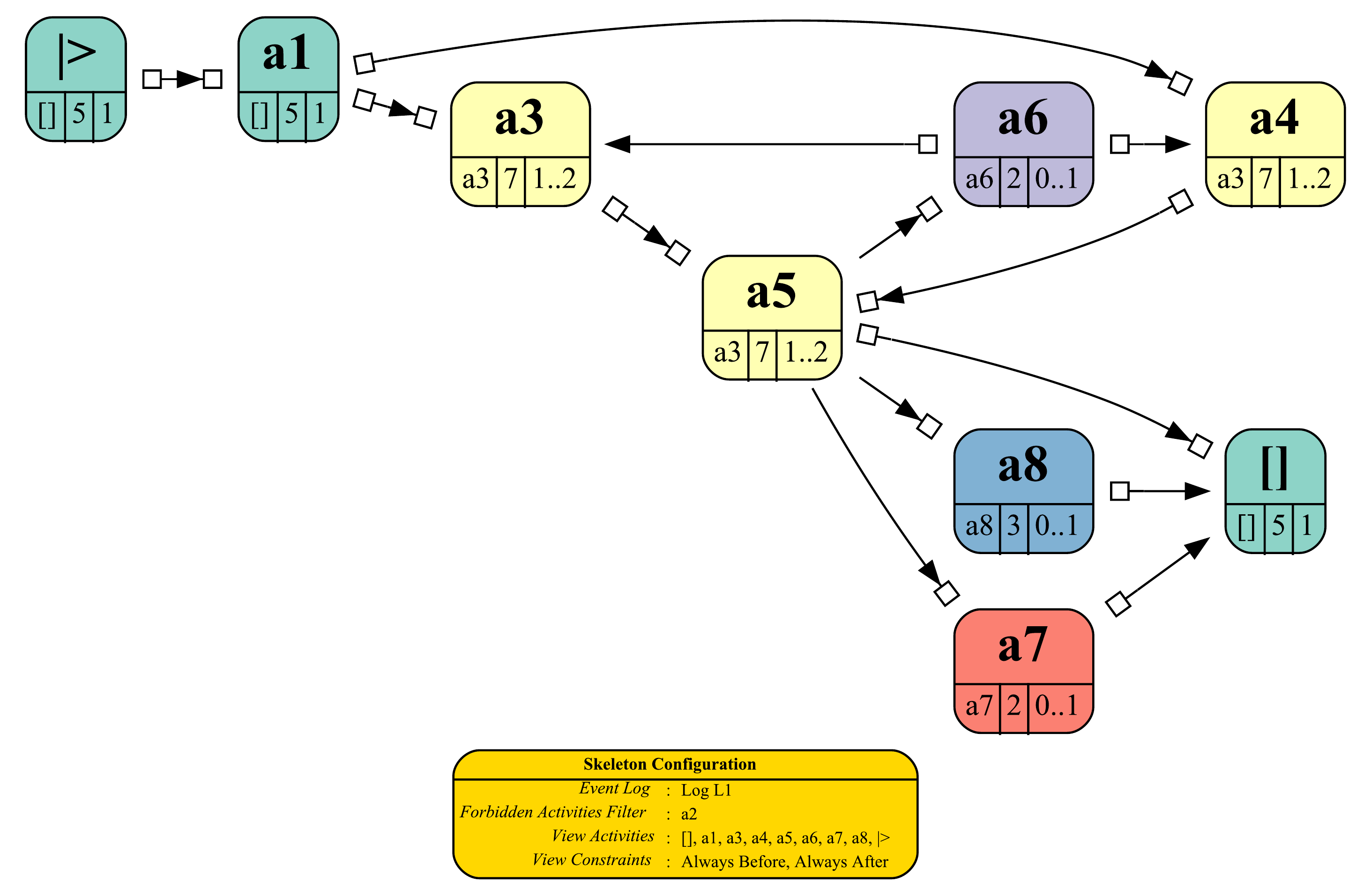}
\caption{Default log skeleton for activity log $L_2$ ($L_1$ with activity $a_2$ forbidden).}
\label{fig:skltn:fbdA2}
\end{figure}
Figure~\ref{fig:skltn:fbdA2} shows \replace{the resulting default}{a} log skeleton \replace{in}{for} that case, which shows that $(a_3,a_5) \in \req{L_2}$, where $L_2 = \fltr{L_1}{\eset{}}{\eset{a_2}}$.
However, $(a_3,a_5) \not\in \req{L_3}$, where $L_3 = \fltr{\ebag{\eseq{a_1,a_4,a_5,a_7}}}{\eset{}}{\eset{a_2}}$.
As a result, log $L_1$ does not subsume the trace $\eseq{a_1,a_4,a_5,a_7}$.

Having the subsumption relation in place, the classification becomes simple: If the log subsumes the trace, then the trace is classified as positive, otherwise as negative.
It is straightforward to check that a log always subsumes any trace it contains: $\qntr{\forall}{\sigma \in L}{\mtch{L}{\sigma}}$:
In Definition~\ref{def:sbsmsl} the implication for \replace{all}{every} universal relation ($\req{}$, $\raa{}$, and $\rab{}$) goes from left to right while the implication for the only existential relation ($\rdf{}$) goes from right to left, and from Definition~\ref{def:sbsms} it is clear that after filtering all filtered traces are still contained in the filtered log.
As a result, a trace from the log itself will always be classified positive.

\section{Implementation}\label{sec:impl}

The log skeleton model and its visual representation have been implemented in the ProM 6 package called LogSkeleton\footnote{See \url{https://svn.win.tue.nl/repos/prom/Packages/LogSkeleton/Trunk} for the sources.}.
After having imported an activity log in ProM 6, this log can be visualized by the \emph{Log Skeleton Filter and Browser} plug-in.
\begin{figure}[tb]
\centering
\includegraphics[width=\textwidth]{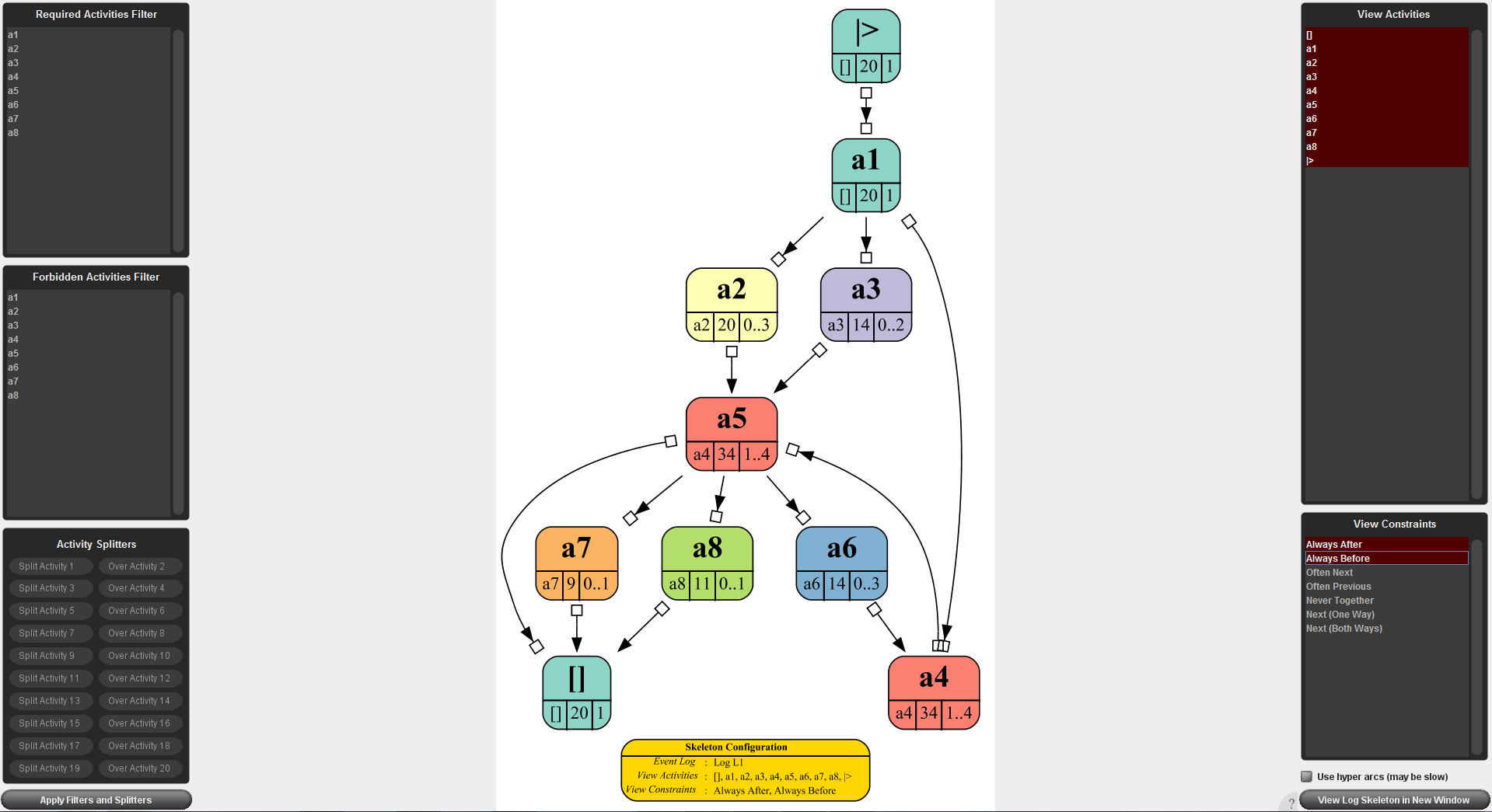}
\caption{The default skeleton visualization for activity log $L_1$.}
\label{fig:skltn:plugin}
\end{figure}
Figure~\ref{fig:skltn:plugin} shows the default skeleton visualization for activity log $L_1$.
In the middle of the visualization, we see the log skeleton that was constructed from the extended log.
The rounded box with yellow background on the bottom shows details for the visualized skeleton, which includes the name of the log it originates from, which activities were selected, which relations were selected, and some more.
This way, all relevant information on how to reconstruct this skeleton and its visualization is at hand.

At the right-hand side, from top to bottom, we see:
\begin{enumerate}
\item A \emph{View Activities} multi-selection box, which allows the user to select which activities (including the artificial activities) to visualize.
By default, all activities are selected.
\item A \emph{View Constraints} multi-selection box, which allows the user to select which relations (or constraints) to visualize.
By default only the \emph{always} relations are selected.
\item A checkbox that allows the user to group edges in the visualized graph into hyper edges.
\item A button that allows the user to visualize the current skeleton with all selections in a separate window.
This may be handy if the user wants to keep the skeleton and selections for later use.
\end{enumerate}
At the left-hand side, from top to bottom, we see (we use $L$ for the original log):
\begin{enumerate}
\item A \emph{Required Activities Filter} multi-selection box, which allows the user to select the set of required activities $A^{\mathrm{req}}$.
As filtering in the artificial activities makes no sense (they are always present in any trace), these artificial activities are not included in this box.
\item A \emph{Forbidden Activities Filter} multi-selection box, which allows the user to select the set of forbidden activities $A^{\mathrm{fbd}}$.
As filtering out the artificial activities makes no sense (they are always present in any trace), these artificial activities are not included in this box.
This filter can be combined with the \emph{Required Activities Filter}, which results in the filtered log $\fltr{L}{A^\mathrm{req}}{A^\mathrm{fbd}}$.
\item An \emph{Activity Splitters} two-column table, which is out-of-scope for this paper.
\item A button that allows the user to construct a new skeleton from the log obtained through the selected filters and splitters.
\end{enumerate}

\section{Process Discovery Contest}\label{sec:contest}

We used the \replace{log skeletons and the plug-in}{implementation} to participate in the Process Discovery Contest (PDC) of 2017.
However, we participated in a way that \replace{}{also }included manual steps.
In the end, this resulted in a 100\% correct classification, that is, all $200$ traces were classified correctly.
In this paper, we restrict ourselves to the fully\replace{ }{-}automated part of the approach, as this makes the comparison to fully-automated discovery algorithms possible.

The PDC of 2017 contained ten test cases $C_1,\ldots,C_{10}$, where every test case $C_i$ consisted of a system $S_i$ and four logs:
A training log $L^0_i$, a first calibration log $L^1_i$, a second calibration log $L^2_i$, and a test log $L^3_i$.
Every training log $L^0_i$ was generated using the system $S_i$, although noise was added for some test cases to these training logs.
Provided only the logs (the systems were not disclosed), the participants had to classify every trace in the test log $L^3_i$ as positive if it could be generated by the system $S_i$, and as negative otherwise.
The participants could use the calibration logs to improve on their approach.

Every training log contains 1000 traces, every calibration log contains 10 positive and 10 negative traces, and every test log also contains 10 positive and 10 negative traces.
5 out of 10 training logs contained noise in 200 out of 1000 traces, where the noise was limited to having truncated traces, that is, 200 traces were incomplete in such noisy training logs.
Furthermore, for every test case it was given which special constructs (loops, duplicates, long-term dependencies, etc.) were used by that test case.

The classification of the calibration and test logs is done in an automated way.
Basically, this classification works by taking a trace $\sigma$ from any of these logs and the corresponding training log $L^0_i$, and to check whether $L^0_i$ subsumes $\sigma$.
If so, then $\sigma$ is classified as a positive trace, otherwise as a negative trace.
However, in certain aspects the implementation deviates a bit from the formalization.

First of all, the implementation does not check any possible set of required and/or forbidden activities (see Definition~\ref{def:sbsms}), as this would take too much time.
Instead, the implementation limits \replace{both }{}the number of required \replace{activities }and \replace{the number of }{}forbidden activities to at most $3$.
In Definition~\ref{def:sbsms}, this means adding the additional \replace{requirements $\card{A^{\mathrm{req}}} \leq 3$ and $\card{A^{\mathrm{fbd}}} \leq 3$)}{requirement $\card{A^{\mathrm{req}}} + \card{A^{\mathrm{fbd}}} \leq 3$}.

\replace{}{Second, if a filtered log does not subsume a filtered trace because of the $\rdf{}$ (directly follows) relation, we want to have some \emph{support} in the filtered log.
In Definition~\ref{def:sbsms}, this means adding the additional requirement that $\card{\fltr{L}{A^{\mathrm{req}}}{A^{\mathrm{fbd}}}} \ge 16$, that is, the filtered log should contain at least 16 traces. }

\replace{Second}{Third}, it is known that each calibration log and each test log contains $10$ positive traces and $10$ negative traces.
As a result, the implementation stops classifying traces as negative if we already have classified at least $10$ traces as negative.
However, we want to classify those 10 traces as negative of which we are most certain.
As an example, we consider a violation of the equivalence relation to be more important than a violation of the directly-follows relation.
For this reason, the implementation checks the relations used for the subsumption in the following order:
\begin{enumerate}
\item The equivalence relation $\req{}$ and the always relations $\raa{}$ and $\rab{}$ on the entire log.
\item The equivalence relation $\req{}$, first with one required or forbidden activity, then with two, and last with three.
\item The always relations $\raa{}$ and $\rab{}$, first with one required or forbidden activity, then with two, and last with three.
\item The directly-follows relation $\rdf{}$, first with one required or forbidden activity, then with two, and last with three.
\end{enumerate}
As soon as $10$ or more traces have been classified as negative, the remaining traces will be classified positive and the implementation stops.
As a result, the directly follows relation is only checked if from the other (stronger) relations no $10$ traces could be classified as negative.

\begin{table*}[tb]
\centering
\caption{Classification results for the PDC of 2017. ``$+$'' denotes a positive classification, ``eq'', ``aa'', ``ab'', and ``df'' denote a negative classification because of the $\req{}$, $\raa{}$, $\rab{}$, and $\rdf{}$ relation (see Definition~\ref{def:sbsmsl}). A shaded (red) background denotes a false classification, no background denotes a true classification.}
\label{tab:res:auto}
\begin{tabular}{|c|lccccccccccr|lccccccccccr|c|}
\toprule
 & & $\sigma_1$ & $\sigma_2$ & $\sigma_3$ & $\sigma_4$  & $\sigma_5$ & $\sigma_6$ & $\sigma_7$ & $\sigma_8$ & $\sigma_9$ & $\sigma_{10}$	&
  &	& $\sigma_{11}$ & $\sigma_{12}$ & $\sigma_{13}$ & $\sigma_{14}$  & $\sigma_{15}$ & $\sigma_{16}$ & $\sigma_{17}$ & $\sigma_{18}$ & $\sigma_{19}$ & $\sigma_{20}$ &	& $\#$true \\\midrule
$C_1$	&	& eq	& ab	& $+$	& ab	& $+$	& eq	& ab	& ab	& eq	& $+$	&	&	& $+$	& $+$	& ab	& $+$	& $+$	& $+$	& $+$	& $+$	& eq	& df &	& 20 \\
$C_2$	&	& \cellcolor{red!25}$+$	& ab	& $+$	& $+$	& $+$	& ab	& $+$	& ab	& ab	& $+$	&	&	& df	& eq	& eq	& $+$	& $+$	& $+$	& eq	& $+$	& $+$	& ab &	& 19 \\
$C_3$	&	& aa	& $+$	& $+$	& $+$	& eq	& $+$	& eq	& ab	& $+$	& $+$	&	&	& aa	& $+$	& $+$	& $+$	& ab	& ab	& $+$	& eq	& eq	& ab &	& 20 \\
$C_4$	&	& df	& $+$	& eq	& $+$	& $+$	& ab	& $+$	& ab	& $+$	&	$+$	& &	& df	& $+$	& eq	& $+$	& aa	& $+$	& \cellcolor{red!25}ab	& eq	& aa	& eq&	& 19 \\
$C_5$	&	& $+$	& eq	& $+$	& eq	& ab	& $+$	& eq	& ab	& $+$	& eq	&	&	& $+$	& $+$	& $+$	& eq	& ab	& eq	& $+$	& eq	& $+$	& $+$ &	& 20 \\
$C_6$	&	& ab	& $+$	& aa	& $+$	& $+$	& eq	& $+$	& $+$	& $+$	& $+$	&	&	& $+$	& ab	& eq	& ab	& eq	& $+$	& $+$	& aa	& ab	& aa &	& 20 \\
$C_7$	&	& $+$	& aa	& $+$	& $+$	& $+$	& aa	& $+$	& $+$	& aa	& $+$	&	&	& aa	& $+$	& $+$	& $+$	& df	& ab	& ab	& aa	& ab	& aa &	& 20 \\
$C_8$	&	& $+$	& $+$	& $+$	& ab	& eq	& $+$	& eq	& $+$	& ab	& eq	&	&	& ab	& $+$	& $+$	& $+$	& $+$	& $+$	& ab	& eq	& eq	& ab &	& 20 \\
$C_9$	&	& eq	& eq	& ab	& $+$	& $+$	& eq	& eq	& $+$	& eq	& eq	&	&	& eq	& $+$	& ab	& $+$	& $+$	& ab	& $+$	& $+$	& $+$	& $+$ &	& 20 \\
$C_{10}$	&	& eq	& aa	& df	& \cellcolor{red!25}$+$	& \cellcolor{red!25}$+$	& ab	& $+$	& ab	& \cellcolor{red!25}$+$	& ab	&	&	& $+$	& $+$	& ab	& $+$	& \cellcolor{red!25}ab	& $+$	& $+$	& $+$	& $+$	& $+$ &	& 16 \\\bottomrule
\end{tabular}
\end{table*}
Table~\ref{tab:res:auto} shows the result of the classification: $194$ out of $200$ traces were classified correctly!
The two false negatives are a result of filtering both logs and apparent incompleteness of both logs.
For trace $\sigma_{17}$ of case $C_4$ the activities $a$ and $f$ are forbidden, after which a log containing $113$ traces remains.
In this filtered log, activity $r$ is always before activity $w$, that is, $(w,r) \in \rab{}$.
However, in trace $\sigma_{17} = \eseq{k,o,b,t,w,m,r,u,l,h,n,t,s,i}$ this is not the case, as the only $r$ occurs \emph{after} the only $w$.
Similarly, for trace $\sigma_{15}$ of case $C_{10}$ the activity $a$ is forbidden, after which a log containing $112$ traces remains.
In this filtered log, activity $b$ is always before activity $f$, that is, $(f,b) \in \rab{}$.
However, in trace $\sigma_{15} = \eseq{u,q,e,i,p,o,j,f,b,d}$ this is not the case, as the only $b$ occurs \emph{after} the only $f$.
As we feel that the $113$ and $112$ traces are sufficient to conclude these after-before relations from, we conclude that a perfect set of training logs should have contained traces \replace{(}{}like $\sigma_{17}$ and $\sigma_{15}$\replace{)}{,} which would have prevented the discovery of these, apparently false, relations.

The four false positives originate from the fact that we fail to detect them as negatives.
Apparently, the discovered log skeletons abstract too much from the logs or are hindered too much by the noise to be able to conclude that these traces are indeed negative.
As an example, consider the trace $\sigma_4 = \eseq{a,e,i,q,p,o,j,q,b,q,i,o,g}$ of case $C_{10}$.
Because of the noise in this log, we cannot detect that activities $b$ and $d$ are equivalent, that is,  $(b,d) \not\in \req{L^0_{10}}$.
If we would remove the noise from this log, which would correspond to removing the traces that do not contain a $d$, then $b$ and $d$ become equivalent and then trace $\sigma_4$ would be classified as a negative.
However, removing this noise would have been a manual step, and would not have fitted the fully-automated approach.

The results of the participating fully-automated discovery algorithms ranged between $139$ and $153$, where the Inductive Miner (which is the current de facto fully-automated process discovery algorithm) scored $147$.
This clearly shows that our fully-automated approach outperforms all participating fully-automated discovery approaches.
The results also \replace{shows}{show} that in particular the case $C_{10}$ is a problem for our approach, as of the $6$ false classifications, $4$ were from this case.
\begin{figure}[tb]
\centering
\includegraphics[width=0.990\textwidth,trim={0px 80px 0px 10px},clip]{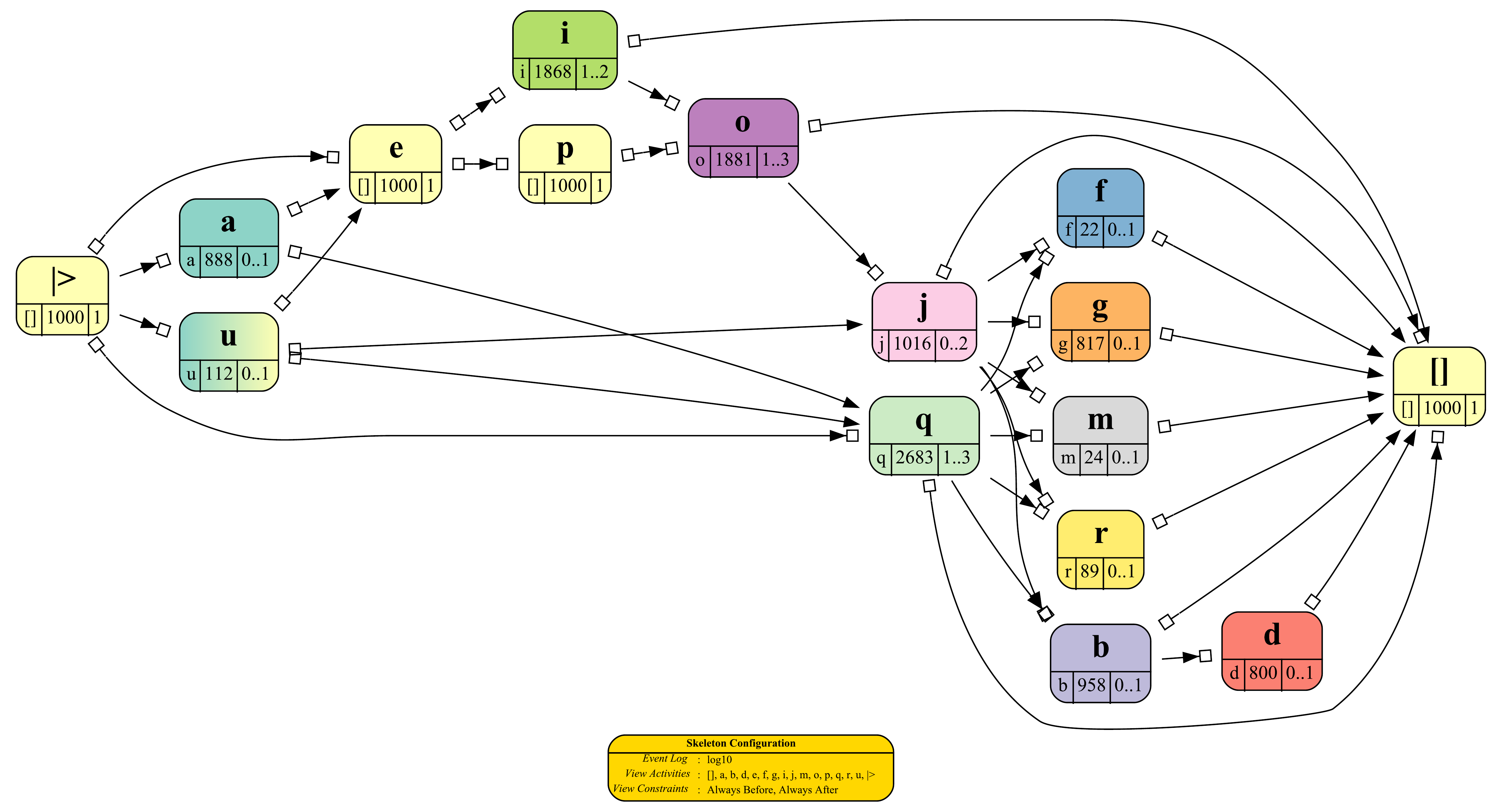}
\caption{Default log skeleton for the log $L^0_{10}$.}
\label{fig:skltn:log10}
\end{figure}
Figure~\ref{fig:skltn:log10} shows the default log skeleton for the corresponding training log.

\section{Conclusions} \label{sec:conc}

This paper has introduced an approach to classify event logs.
Given an event log and a trace, this approach provides a fairly good classification whether or not this trace could have been in this log.
Where using current state-of-the-art fully-automated discovery algorithms like the Inductive Miner~\cite{im} allows one to classify $147$ traces out of $200$ correctly, our fully-automated approach allows one to classify $194$ out of $200$ correctly.

For this classification, our approach uses log skeletons as models.
A log skeleton contains information on the structure in the log using a number of relations, like an \emph{equivalence} relation and \emph{always-after} and \emph{always-before} relations.
\replace{Surprisingly, some}{Some} of these relations correspond to existing Declare~\cite{declare} constraints: The \emph{always-after} relation corresponds to the \emph{succession} constraint and the \emph{always-before} relation to the \emph{precedence} constraint.
Nevertheless, the \emph{equivalence} relation has \replace{not}{no} direct counterpart in Declare, although it implies the \emph{co-existence} relation.
As such, there are strong links between the log skeletons and Declare.
Another difference with Declare is that the classification check for log skeletons also includes filtering the log: A trace conforms to an event log if the skeleton of the log subsumes the skeleton of the trace, and if this also holds for the trace and the log after some activities have been filtered in or out.

As classification is used by the Process Discovery Contest (PDC)~\cite{pdc} to measure the effectiveness of discovery algorithms, this begs the question whether our construction algorithm of log skeletons is not a \emph{very effective discovery algorithm}.
So far, the classification using these log skeleton models classify much better than the models as discovered by existing fully-automated discovery algorithms.
And, on top of that, the construction algorithm for log skeleton works reasonably fast, fast enough to implement if in ProM 6~\cite{prom6} as an event log visualizer.

We participated with the log skeleton approach to the PDC of 2017, but not with the fully-automated variant.
Instead, we used a variant that included manually configured preprocessors for every event log.
These preprocessors included requiring and forbidding activities, and the splitting of so-called duplicate activities.
In the end, these preprocessors improved the classification to a perfect classification for this Contest: all $200$ traces were classified correctly.
To add these preprocessors to the fully-automated variant requires the fully-automated deduction of these filters and/or splitters from the event log at hand.
In the future, we hope to be able to add such a feature to the fully-automated variant.

Other future work includes the conversion from a log skeleton model to a more \emph{mainstream} model like a BPMN (Business Process Model and Notation)~\cite{bpmn} model or a Petri net~\cite{petrinets}.
Although our classification was the best, in the end, we did not win the PDC of 2017 because the jurors \replace{thought}{considered} the BPMN models of a competitor to be more insightful.

Our log skeleton approach is sensitive to noise:
One missing activity might break an always-before and/or an always-after relation, and one missing or one spurious activity might break an equivalence relation.
For the PDC of 2017, the effect of noise was fortunately limited, as only the last part of the trace might be missing.
As a result, the always-before relation was still dependable.
However, for arbitrary logs, noise may be a problem.
A possibility could be to introduce near-equivalence relation and near-always relations, but such relations might break the nice property we now have that traces contained in the log itself will always be classified positive.
Another idea could be to actually use log skeletons to filter out noise:
If by removing a trace from the log, the equivalence relation or an always relation improves, while all other relations do not get worse, then this trace might indeed contain noise.
For example, we could check whether the equivalence relation gets more coarse if we would remove a trace.

Finally, a possible explanation why the classification using log skeletons works way better than the classification using procedural models, like BPMN models and Petri nets, is that there is a strong bias within the discovery community for using these procedural models.
As a result, the organizers of the PDC may also be biased towards these models, which may result in training logs and test logs that are constructed to be especially difficult for these models, but not for models like the log skeletons.
If so, the presented fully-automated approach using log skeletons can be used by the organizers of the PDC to improve on this.

\section*{Acknowledgements}
The authors would like to thank the organizers of the Process Discovery Contest series for their work on these contests.
Without these contests, this work would not have existed.

\bibliographystyle{splncs03}
\bibliography{hverbeek}

\begin{thebibliography}{10}
\providecommand{\url}[1]{\texttt{#1}}
\providecommand{\urlprefix}{URL }

\bibitem{alpha}
Aalst, W.M.P.v.d., Weijters, A.J.M.M., Maruster, L.: Workflow mining:
  Discovering process models from event logs. IEEE Transactions on Knowledge
  and Data Engineering  16(9),  1128--1142 (2004)

\bibitem{wfnets}
Aalst, W.M.P.v.d.: The application of petri nets to workflow management. The
  Journal of Circuits, Systems and Computers  8(1),  21--66 (1998)

\bibitem{pmbook2}
Aalst, W.M.P.v.d.: Process Mining: Data Science in Action (2016)

\bibitem{declare}
Aalst, W.M.P.v.d., Pesic, M., Schonenberg, H.: Declarative workflows: Balancing
  between flexibility and support. Computer Science - Research and Development
  23,  99--113 (2009)

\bibitem{pdc}
Carmona, J., de~Leoni, M., Depaire, B., Jouck, T.: Process discovery contest
  (2016),
  \url{http://www.win.tue.nl/ieeetfpm/doku.php?id=shared:process_discovery_contest}

\bibitem{petrinets}
Desel, J., Reisig, W., Rozenberg, G. (eds.): Lectures on Concurrency and Petri
  Nets, Lecture Notes in Computer Science, vol. 3098. Springer-Verlag, Berlin
  (2004)

\bibitem{im}
Leemans, S.J.J., Fahland, D., Aalst, W.M.P.v.d.: Discovering block-structured
  process models from event logs - a constructive approach. In: Colom, J.M.,
  Desel, J. (eds.) Application and Theory of {Petri} Nets and Concurrency,
  Lecture Notes in Computer Science, vol. 7927, pp. 311--329. Springer Berlin
  Heidelberg (2013), \url{http://dx.doi.org/10.1007/978-3-642-38697-8_17}

\bibitem{prom6}
Verbeek, H.M.W., Buijs, J.C.A.M., Dongen, B.F.v., Aalst, W.M.P.v.d.: {ProM} 6:
  The process mining toolkit. In: Proc. of BPM Demonstration Track 2010. vol.
  615, pp. 34--39. CEUR-WS.org (2010),
  \url{http://ceur-ws.org/Vol-615/paper13.pdf}

\bibitem{heuristics}
Weijters, A.J.M.M., Aalst, W.M.P.v.d.: Rediscovering workflow models from
  event-based data using {Little Thumb}. Integr. Comput.-Aided Eng.  10(2),
  151--162 (Apr 2003), \url{http://dl.acm.org/citation.cfm?id=1273320.1273325}

\bibitem{ilp}
Werf, J.M.E.M.v.d., Dongen, B.F.v., Hurkens, C.A.J., Serebrenik, A.: Process
  discovery using integer linear programming. Fundam. Inf.  94(3-4),  387--412
  (Aug 2009), \url{http://dl.acm.org/citation.cfm?id=1662594.1662600}

\bibitem{bpmn}
Weske, M.: Business Process Management: Concepts, Languages, Architectures.
  Springer-Verlag, Berlin (2007)

\bibitem{hybridilp}
Zelst, S.J.v., Dongen, B.F.v., Aalst, W.M.P.v.d.: Ilp-based process discovery
  using hybrid regions. In: Proceedings of the International Workshop on
  Algorithms {\&} Theories for the Analysis of Event Data, {ATAED} 2015,
  Satellite event of the conferences: 36th International Conference on
  Application and Theory of Petri Nets and Concurrency Petri Nets 2015 and 15th
  International Conference on Application of Concurrency to System Design
  {ACSD} 2015, Brussels, Belgium, June 22-23, 2015. pp. 47--61 (2015),
  \url{http://ceur-ws.org/Vol-1371/paper04.pdf}

\end{thebibliography}

\appendix
\raggedbottom
\section{Default log skeletons}\label{appA}

\subsection{Default log skeleton for log $L^0_1$}
\begin{center}
\includegraphics[width=0.979\textwidth,trim={0px 80px 0px 10px},clip]{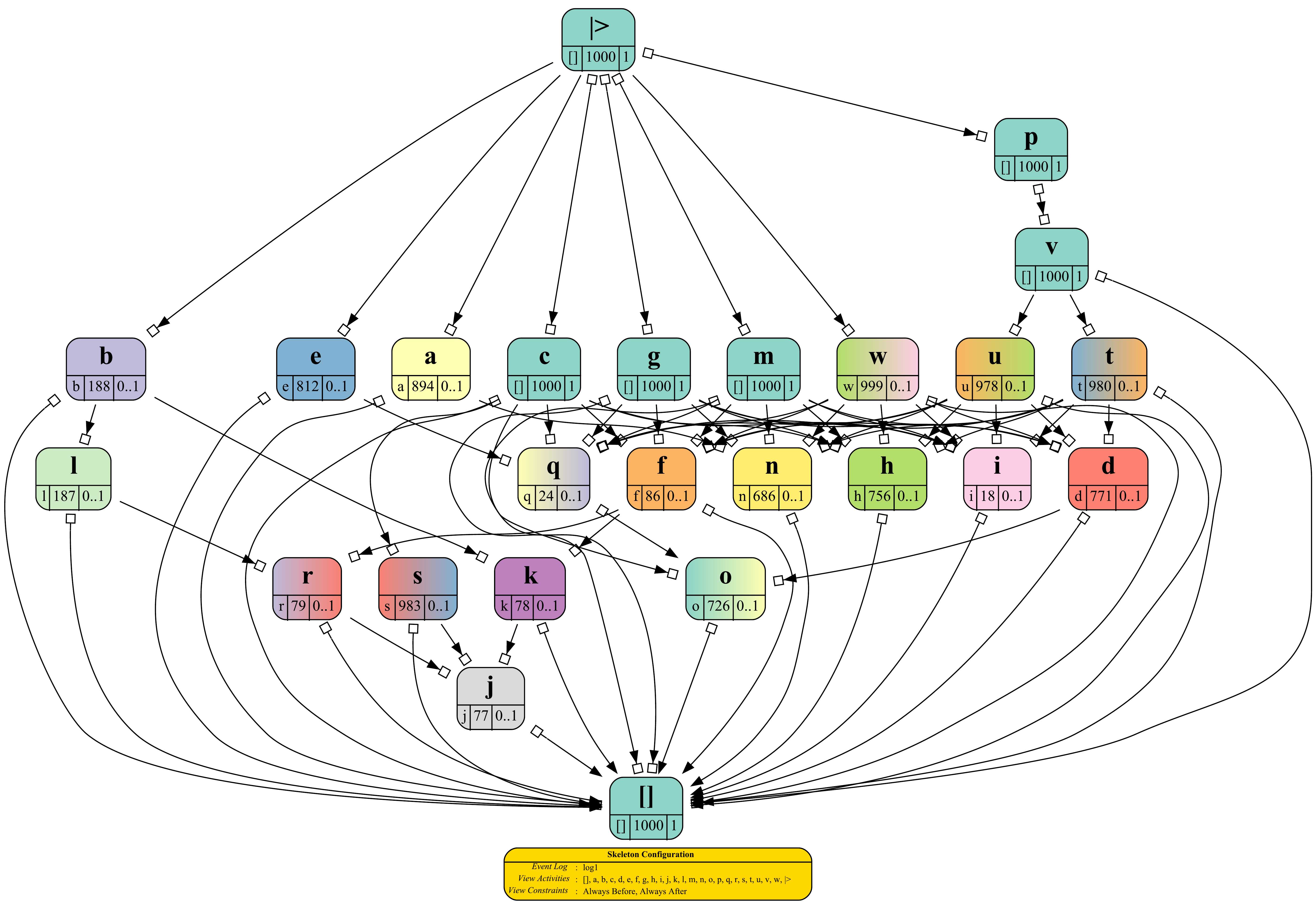}
\end{center}

\subsection{Default log skeleton for log $L^0_2$}
\begin{center}
\includegraphics[width=0.806\textwidth,trim={0px 80px 0px 10px},clip]{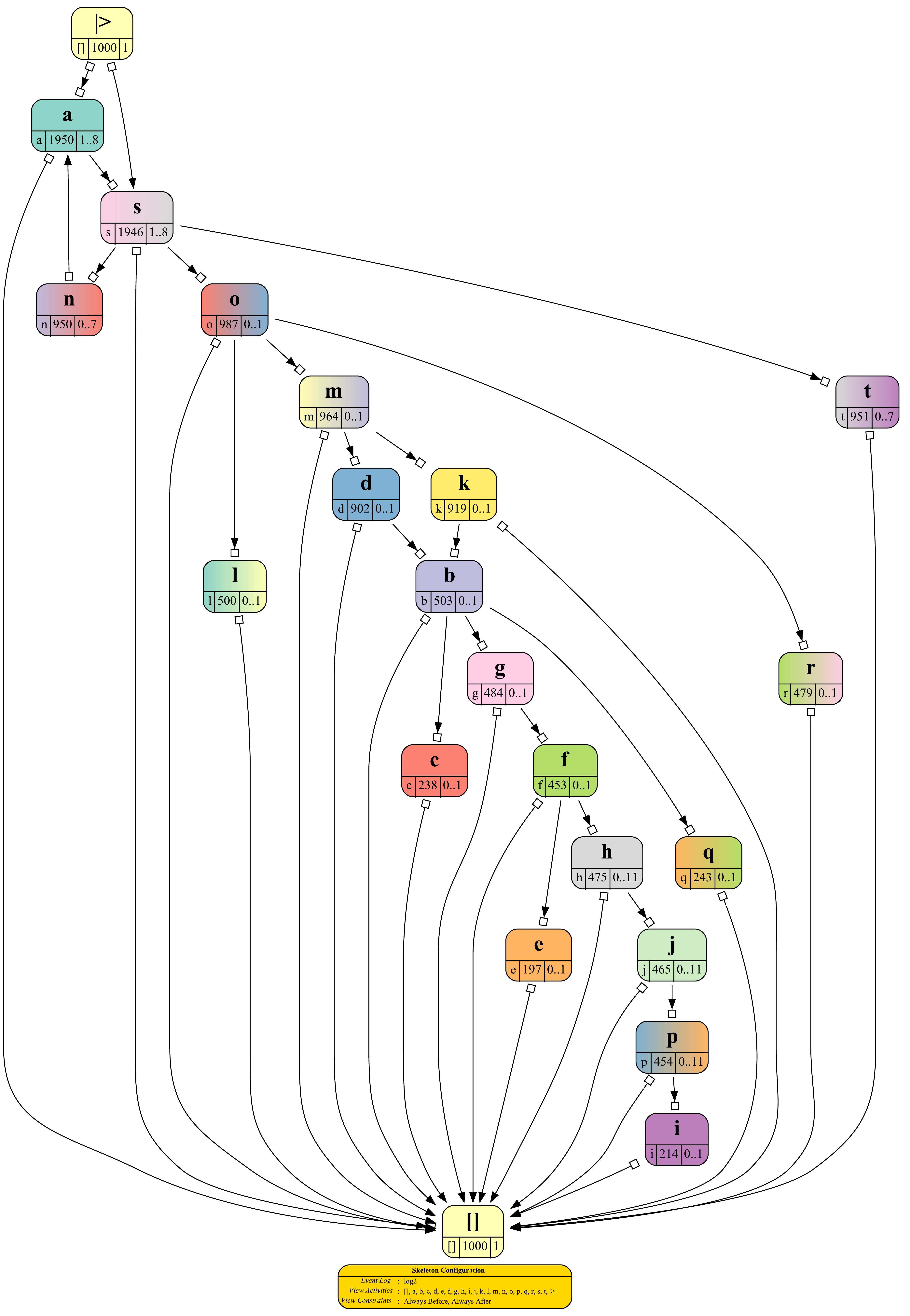}
\end{center}

\subsection{Default log skeleton for log $L^0_3$}
\begin{center}
\includegraphics[width=0.599\textwidth,trim={0px 80px 0px 10px},clip]{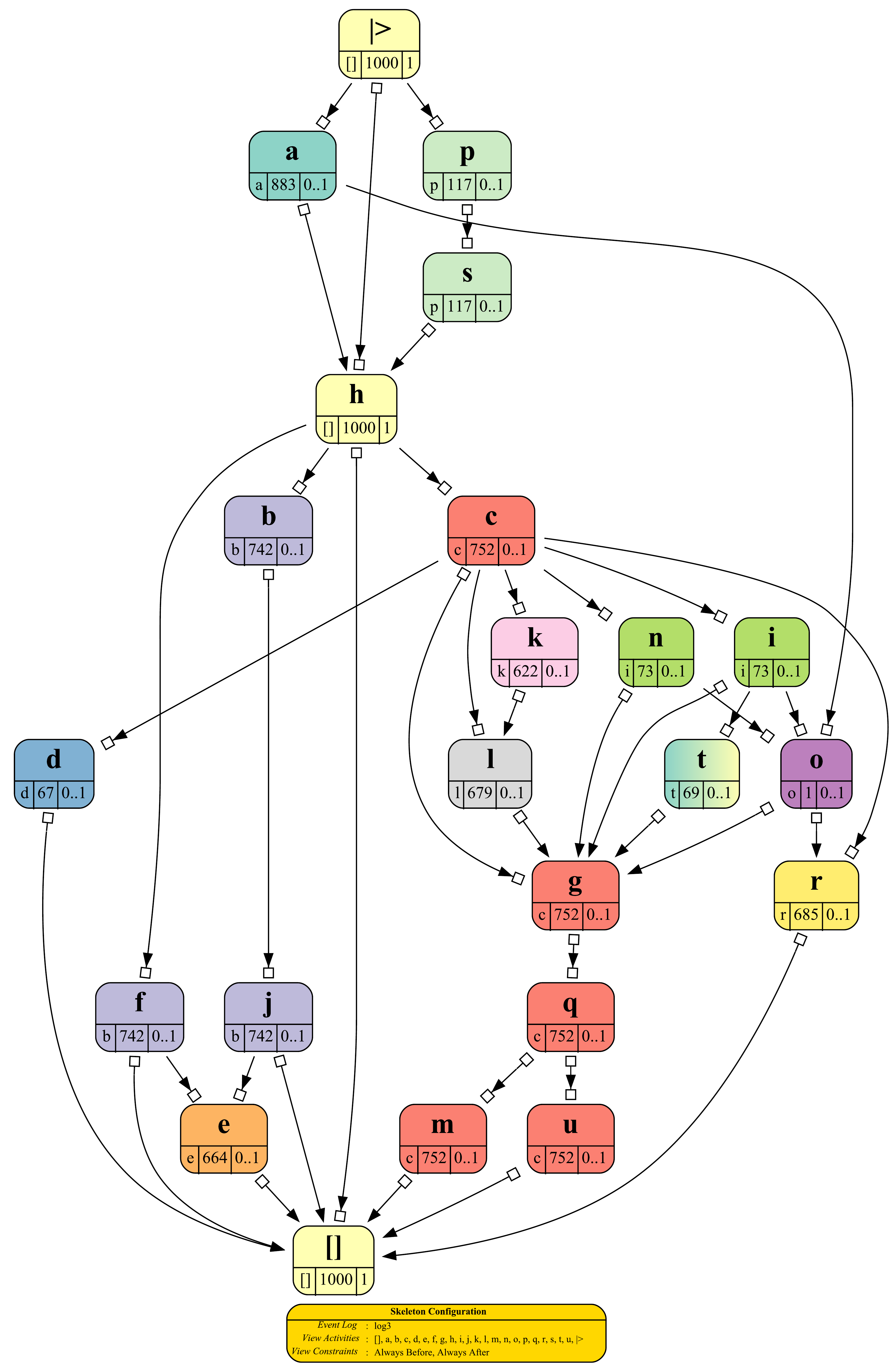}
\end{center}

\subsection{Default log skeleton for log $L^0_4$}
\begin{center}
\includegraphics[width=0.849\textwidth,trim={0px 80px 0px 10px},clip]{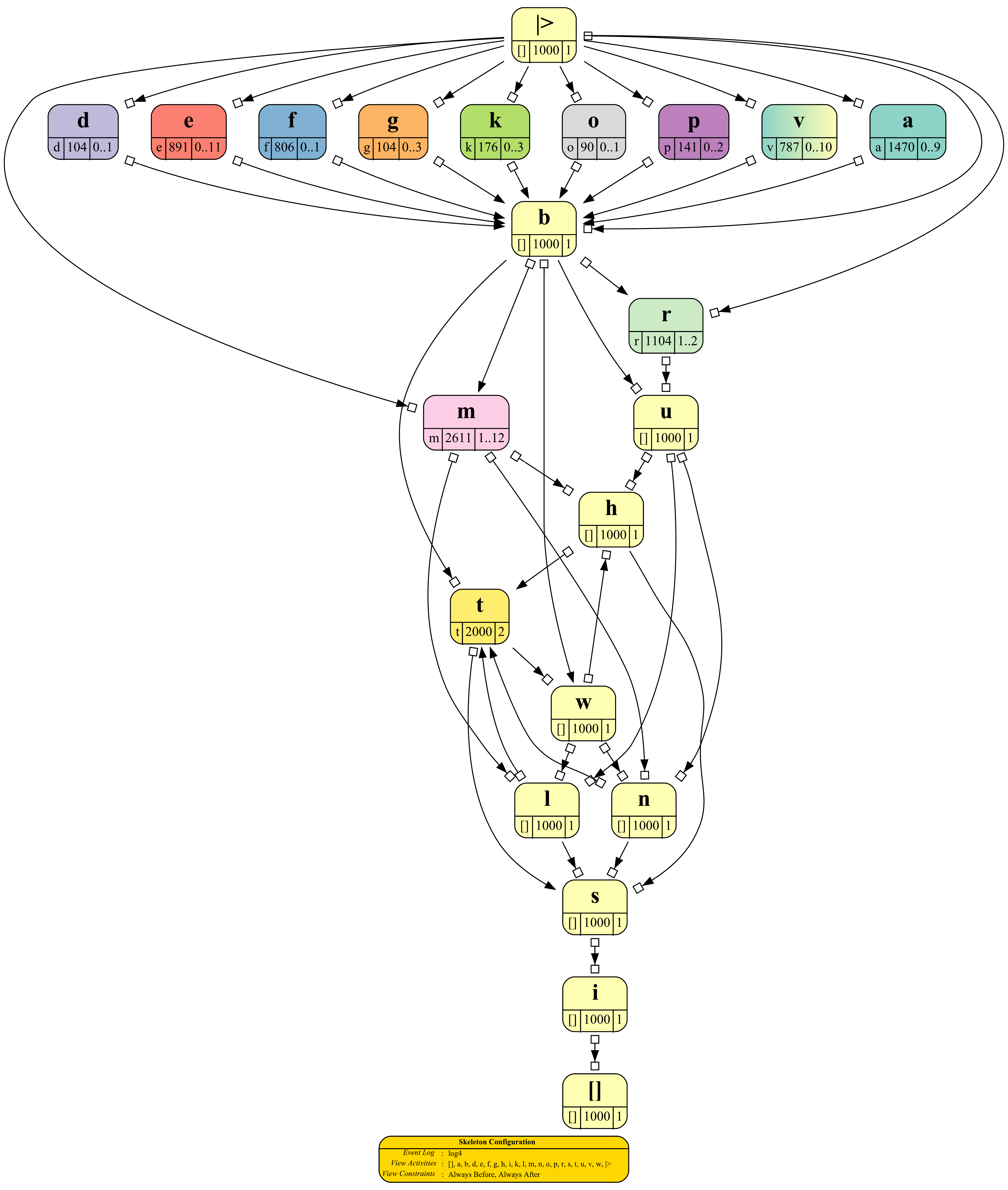}
\end{center}

\subsection{Default log skeleton for log $L^0_5$}
\begin{center}
\includegraphics[width=0.685\textwidth,trim={0px 80px 0px 10px},clip]{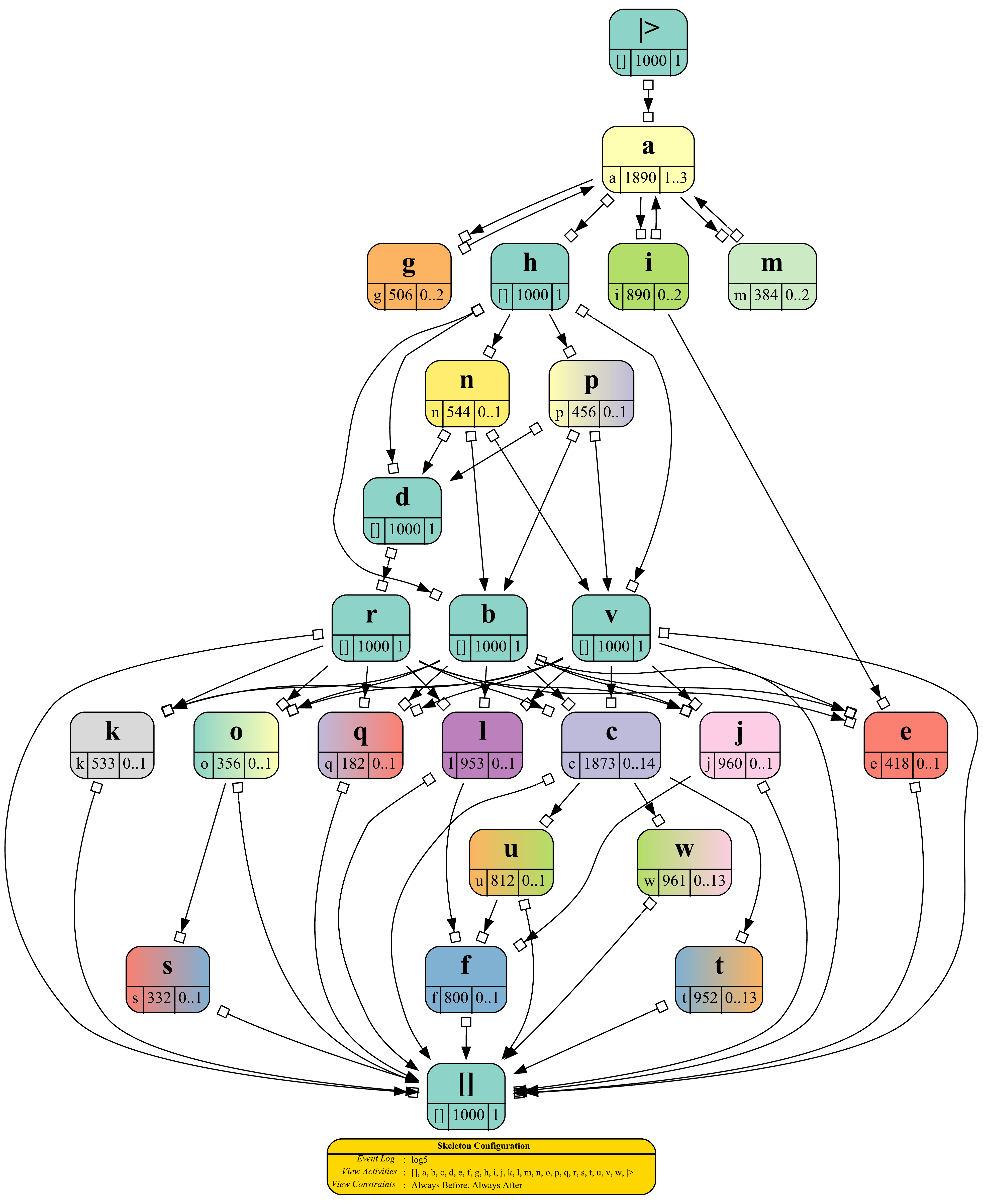}
\end{center}

\subsection{Default log skeleton for log $L^0_6$}
\begin{center}
\includegraphics[width=1.000\textwidth,trim={0px 80px 0px 10px},clip]{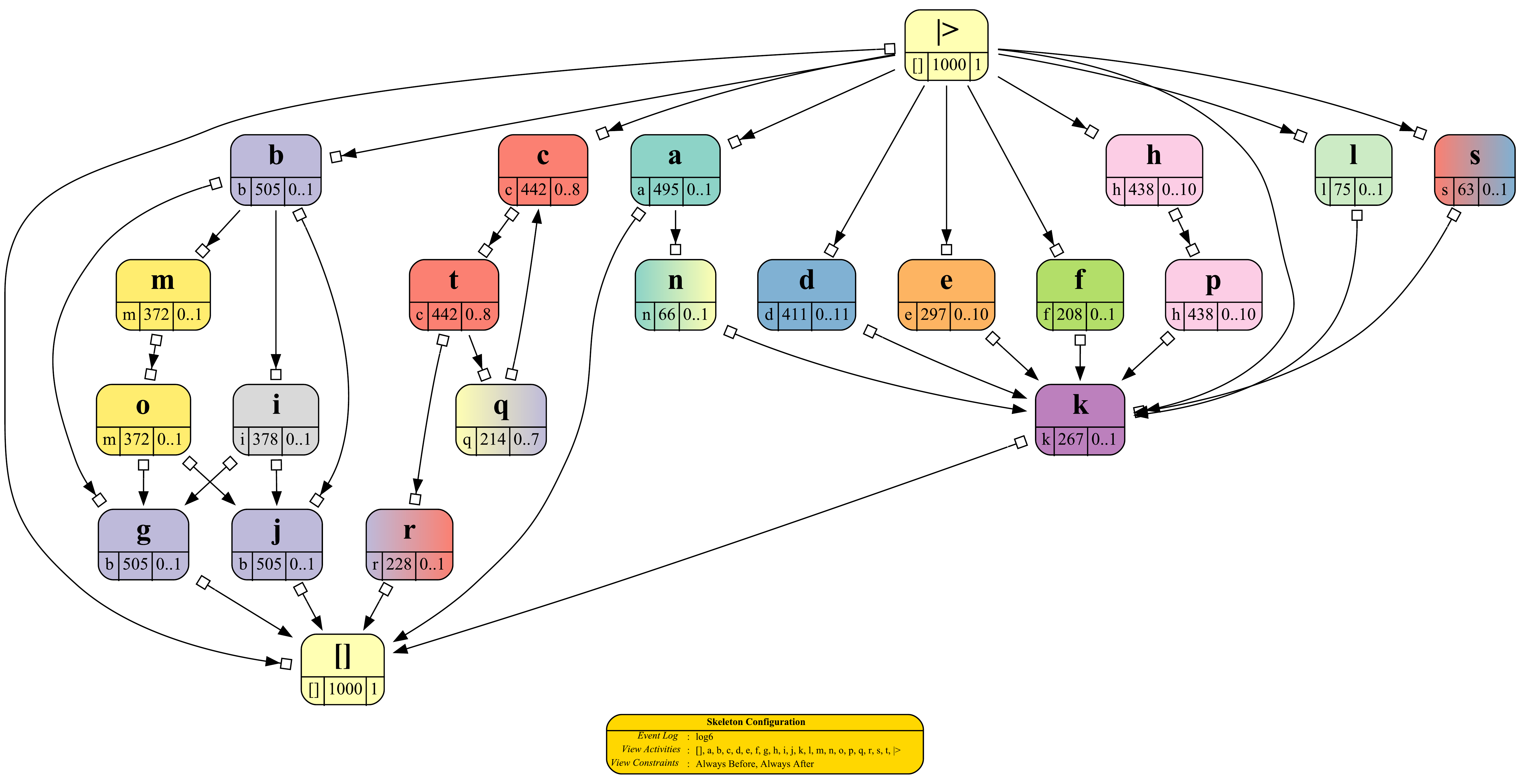}
\end{center}

\subsection{Default log skeleton for log $L^0_7$}
\begin{center}
\includegraphics[width=0.625\textwidth,trim={0px 80px 0px 10px},clip]{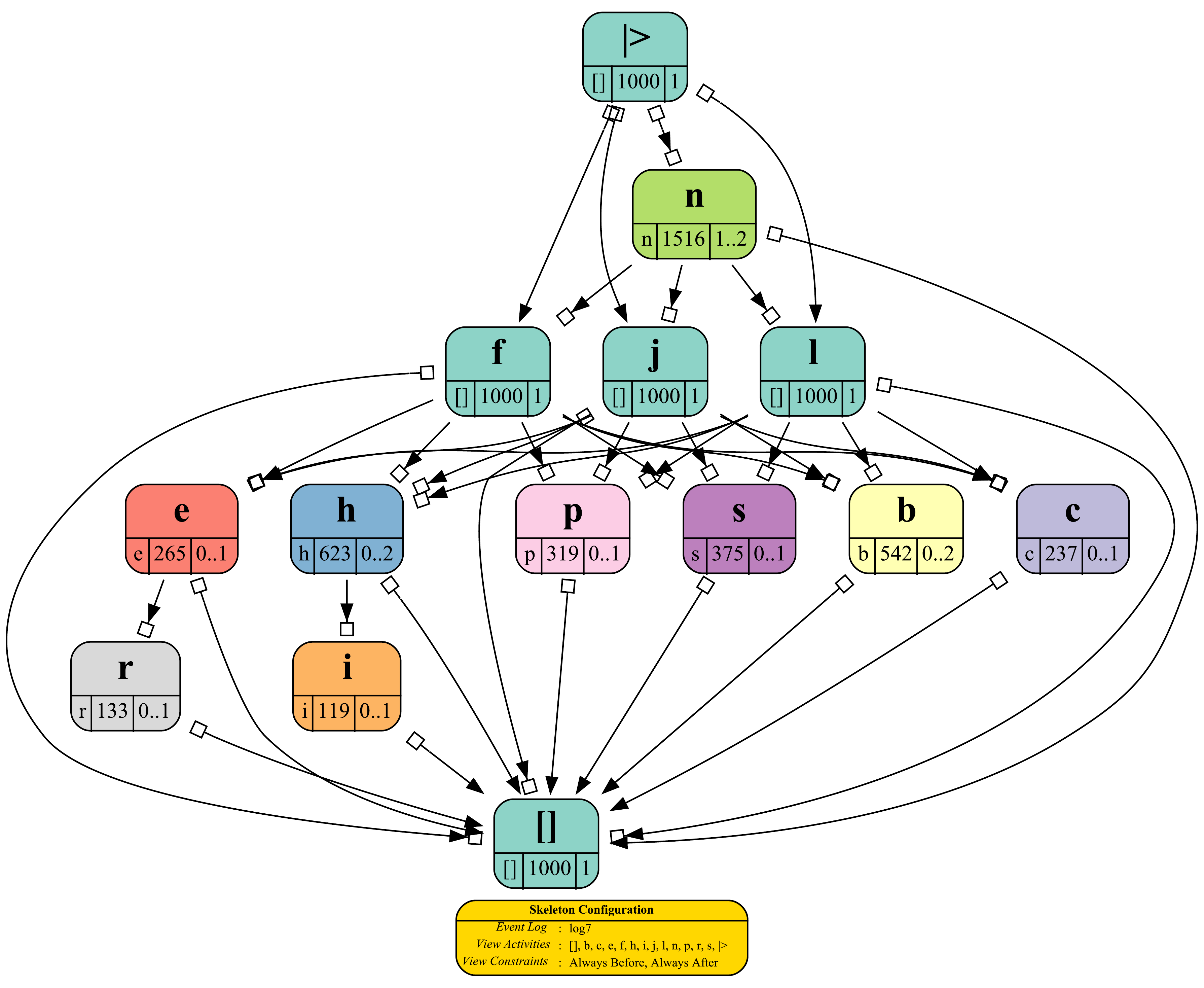}
\end{center}

\subsection{Default log skeleton for log $L^0_8$}
\begin{center}
\includegraphics[width=0.695\textwidth,trim={0px 80px 0px 10px},clip]{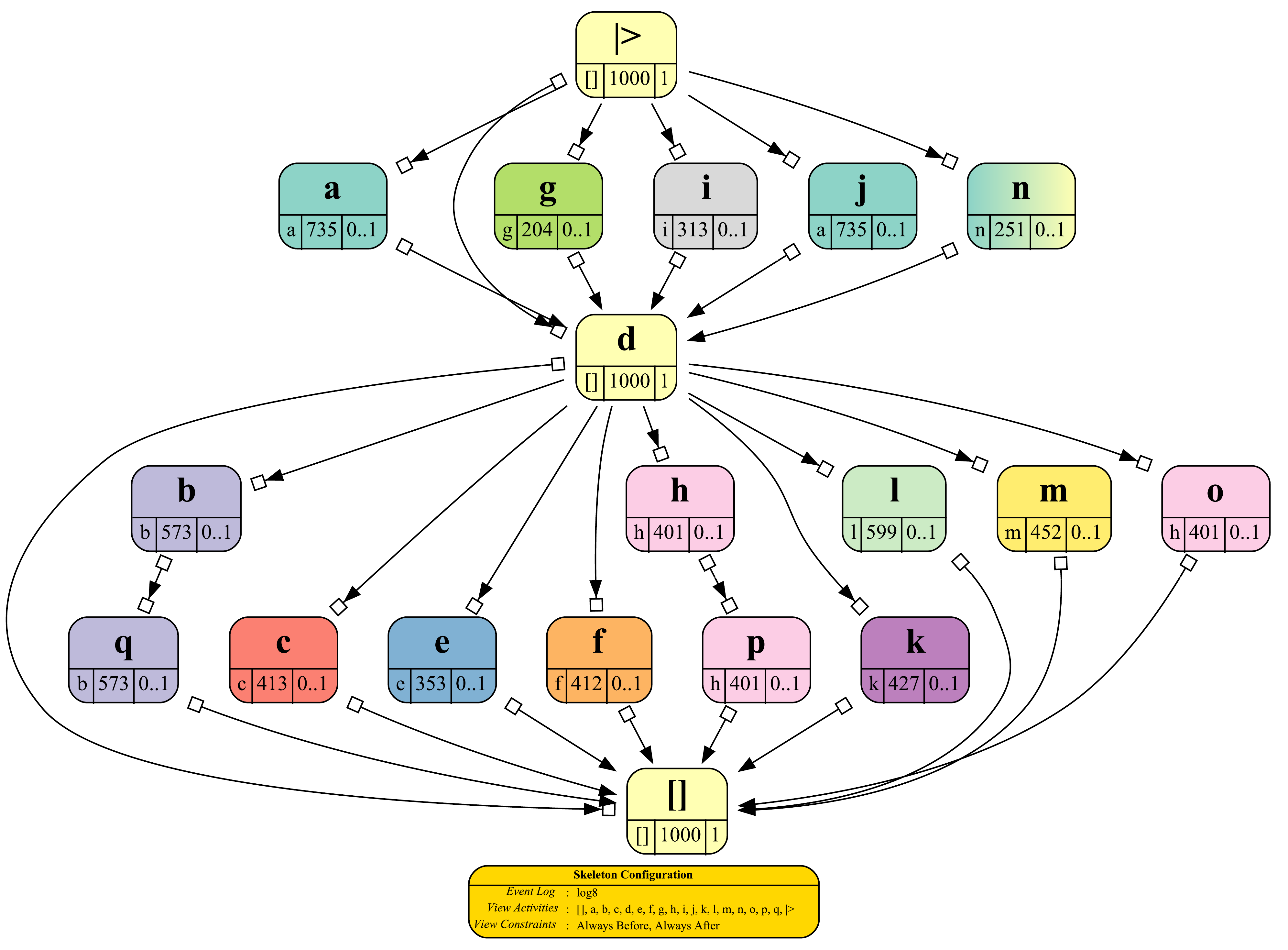}
\end{center}

\subsection{Default log skeleton for log $L^0_9$}
\begin{center}
\includegraphics[width=0.588\textwidth,trim={0px 80px 0px 10px},clip]{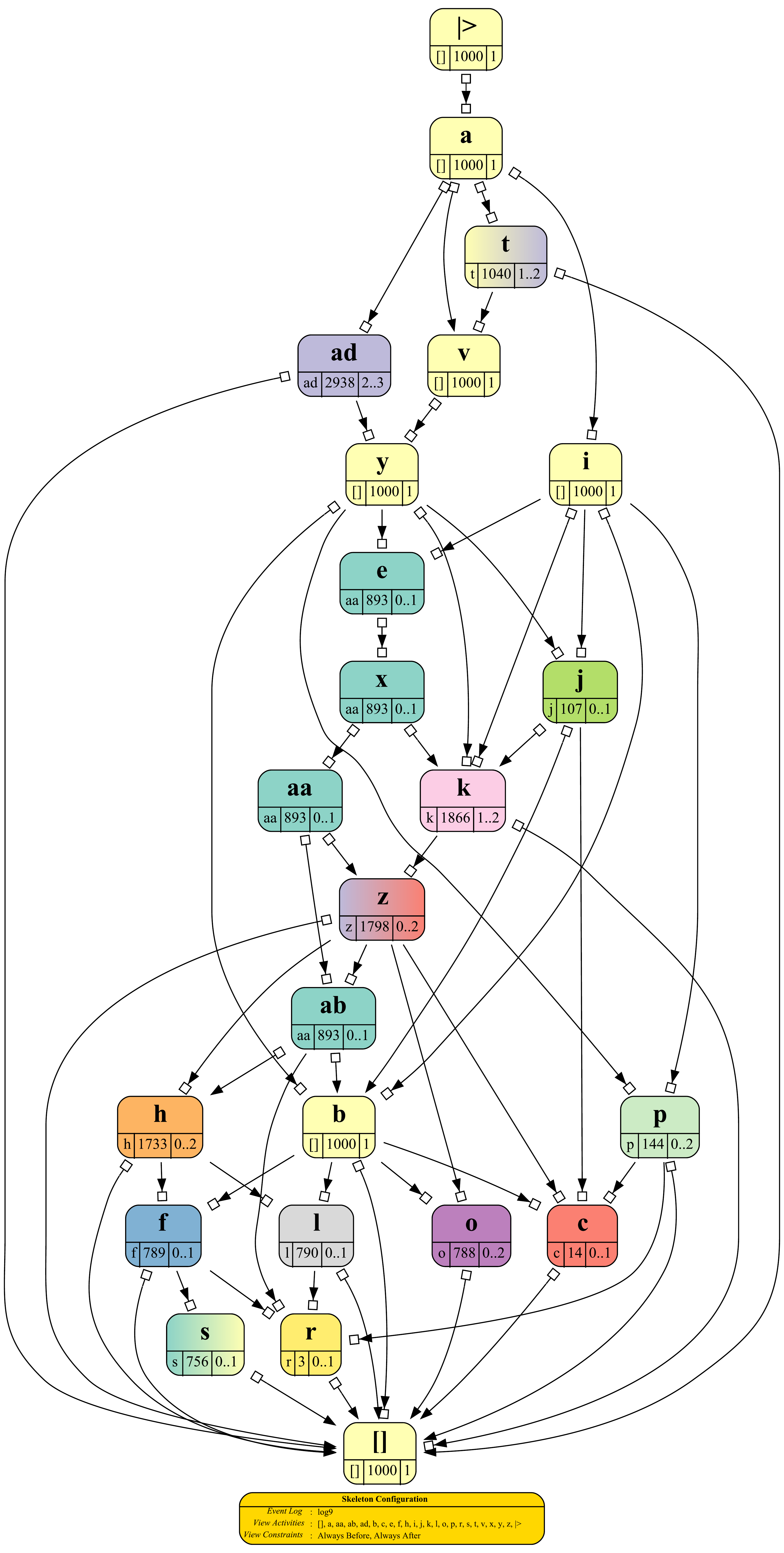}
\end{center}

\subsection{Default log skeleton for log $L^0_{10}$}
\begin{center}
\includegraphics[width=0.518\textwidth,trim={0px 80px 0px 10px},clip]{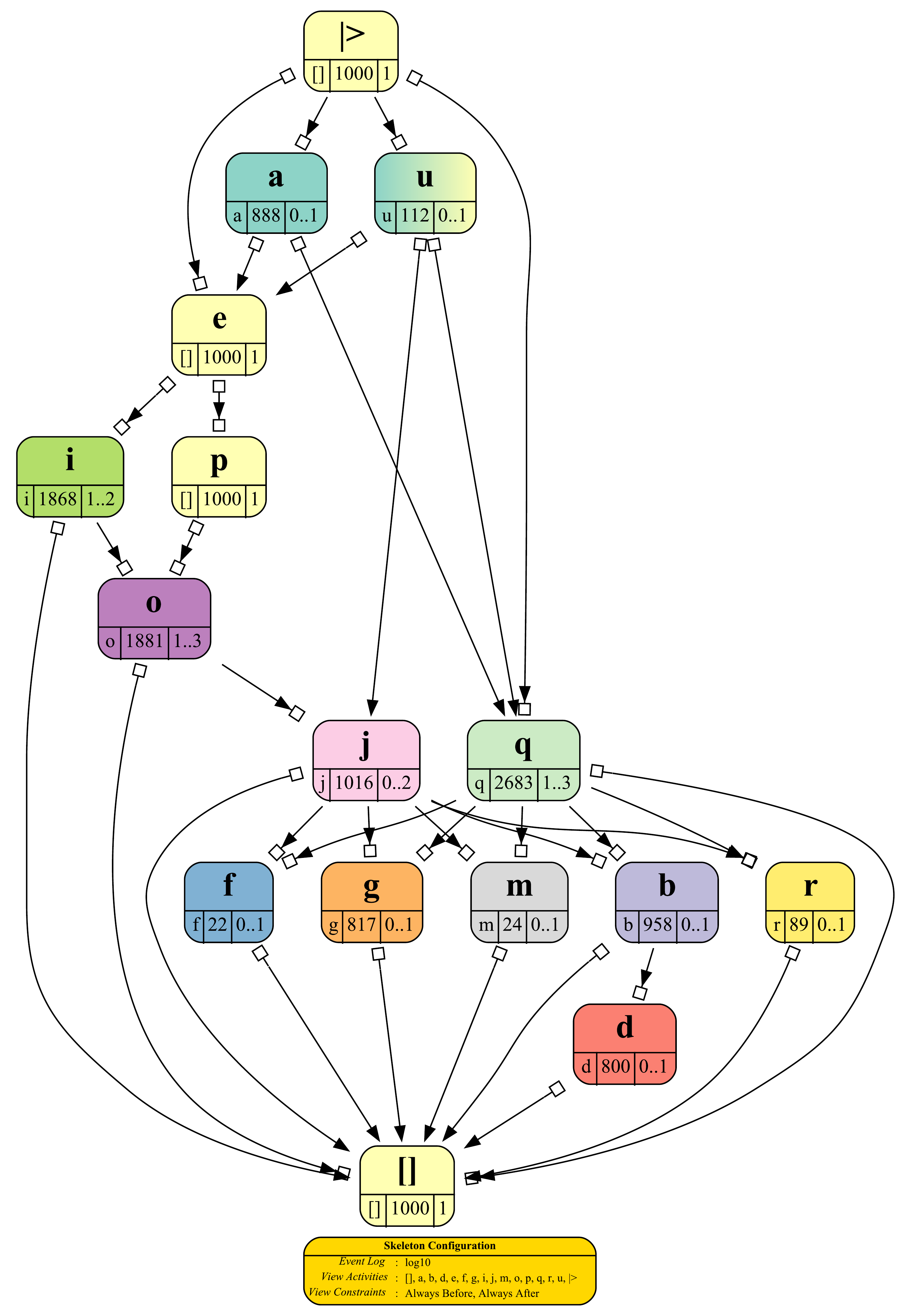}
\end{center}

\section{Log skeletons using hyper arcs}

Some of the log skeletons as shown in Appendix~\ref{appA} show cluttered areas where some source activities have some identical arcs to some target activities.
In this Appendix, we show the same log skeletons but now using hyper arcs, where all identical arcs from multiple source activities to multiple target activities have been replaced by a single hyper arc.

\subsection{Log skeleton with hyper arcs for log $L^0_2$}
\begin{center}
\includegraphics[width=0.969\textwidth,trim={0px 80px 0px 10px},clip]{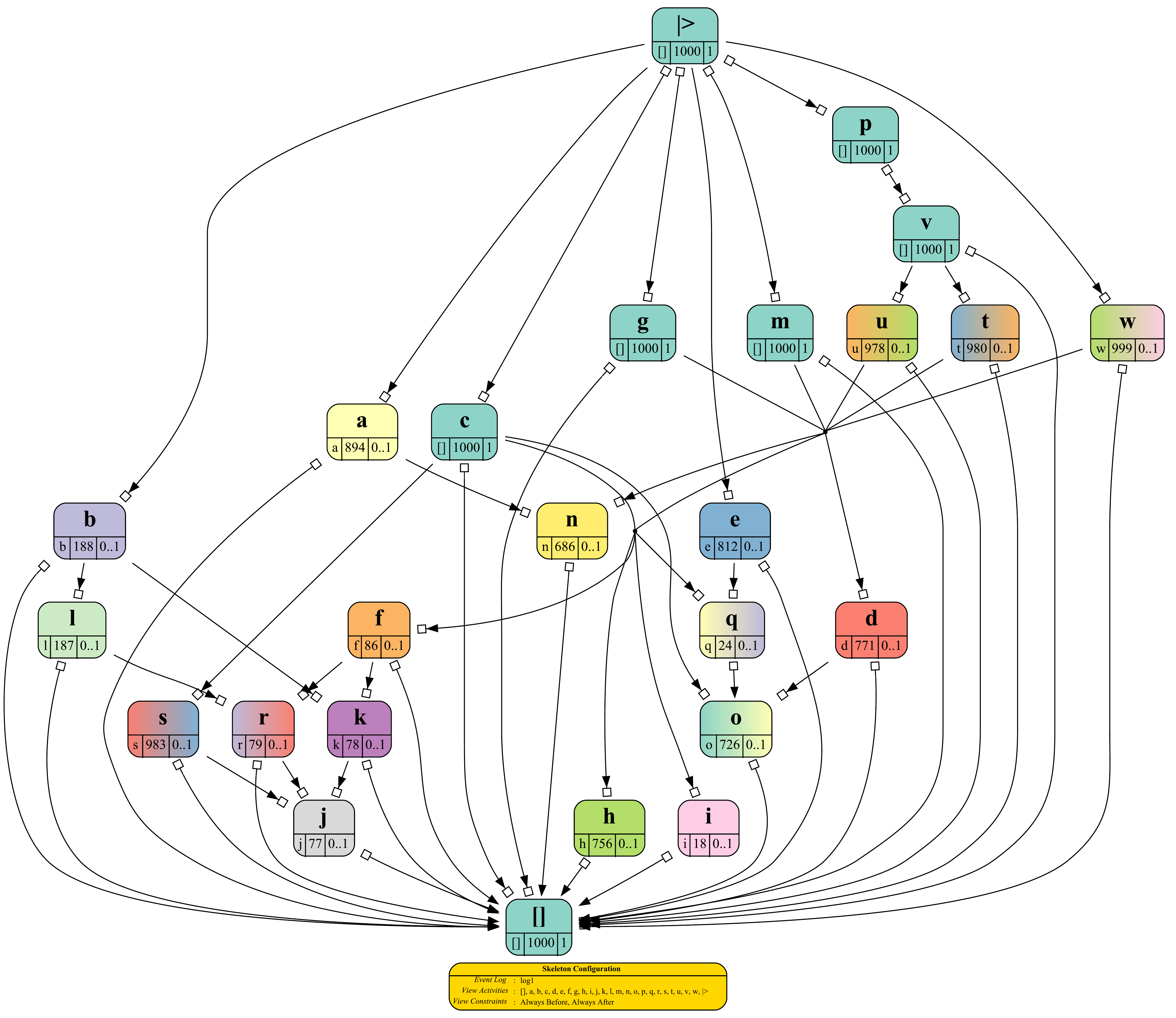}
\end{center}

\subsection{Log skeleton with hyper arcs for log $L^0_4$}
\begin{center}
\includegraphics[width=0.828\textwidth,trim={0px 80px 0px 10px},clip]{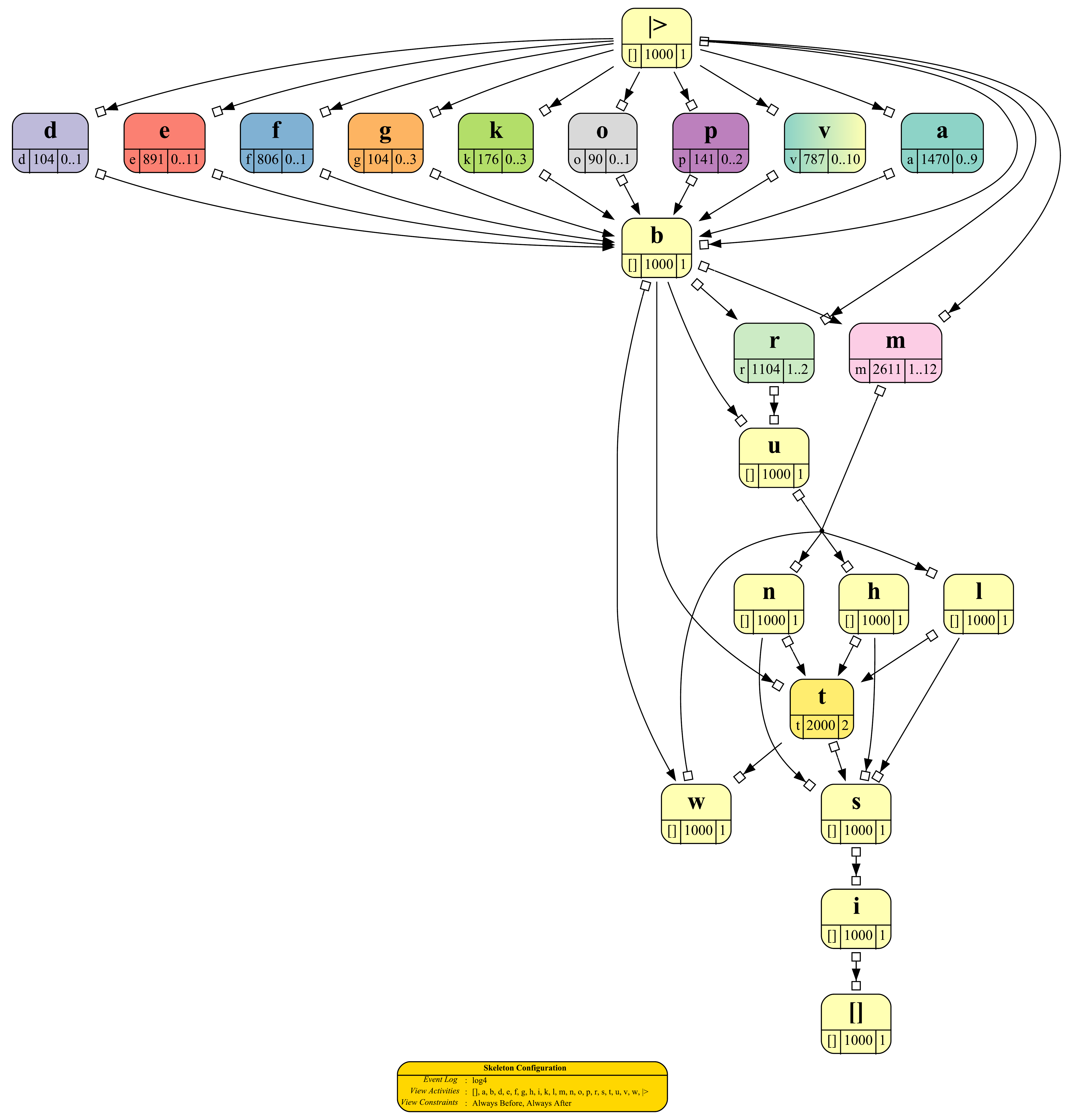}
\end{center}

\subsection{Log skeleton with hyper arcs for log $L^0_5$}
\begin{center}
\includegraphics[width=0.753\textwidth,trim={0px 80px 0px 10px},clip]{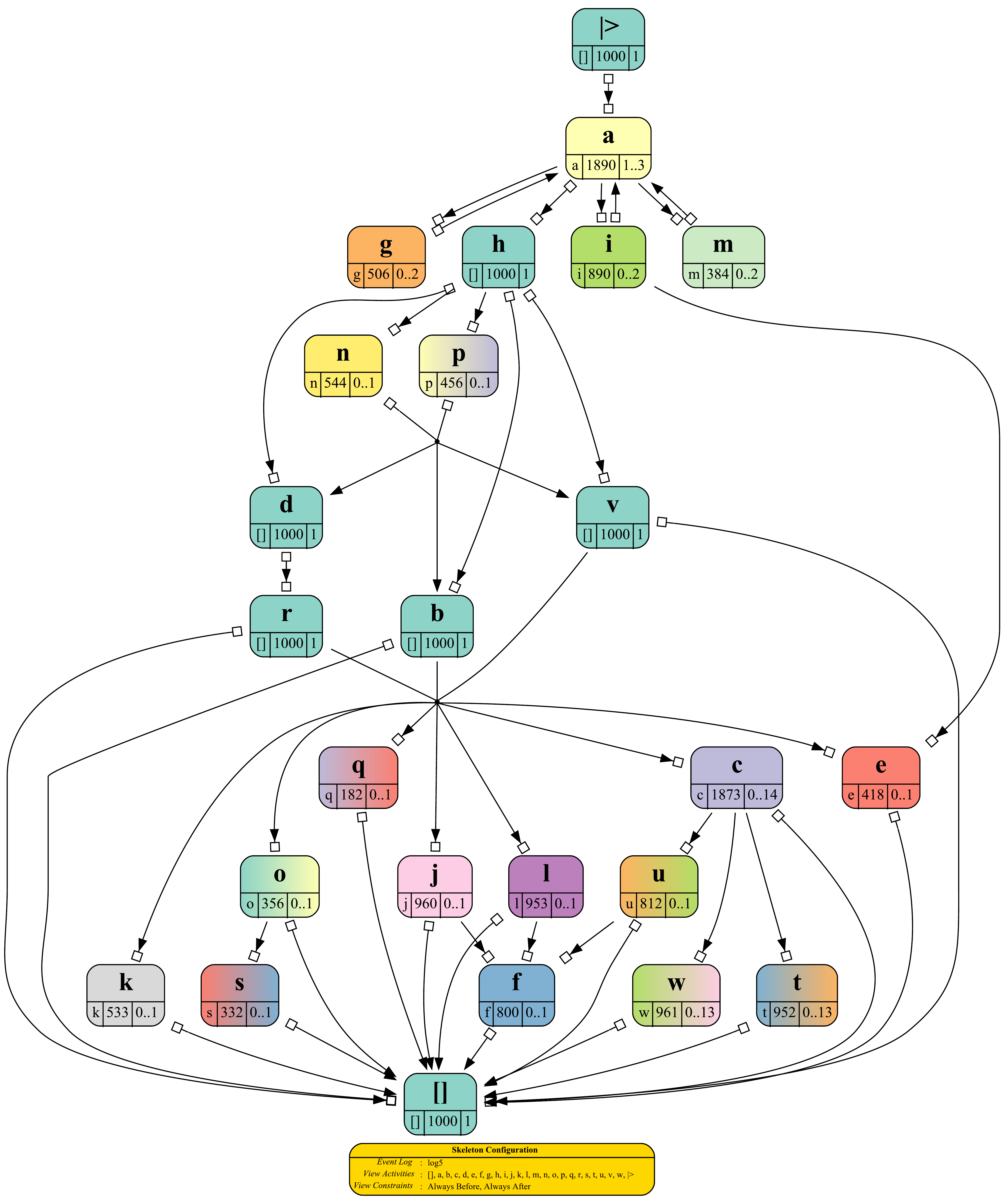}
\end{center}

\subsection{Log skeleton with hyper arcs for log $L^0_6$}
\begin{center}
\includegraphics[width=0.925\textwidth,trim={0px 80px 0px 10px},clip]{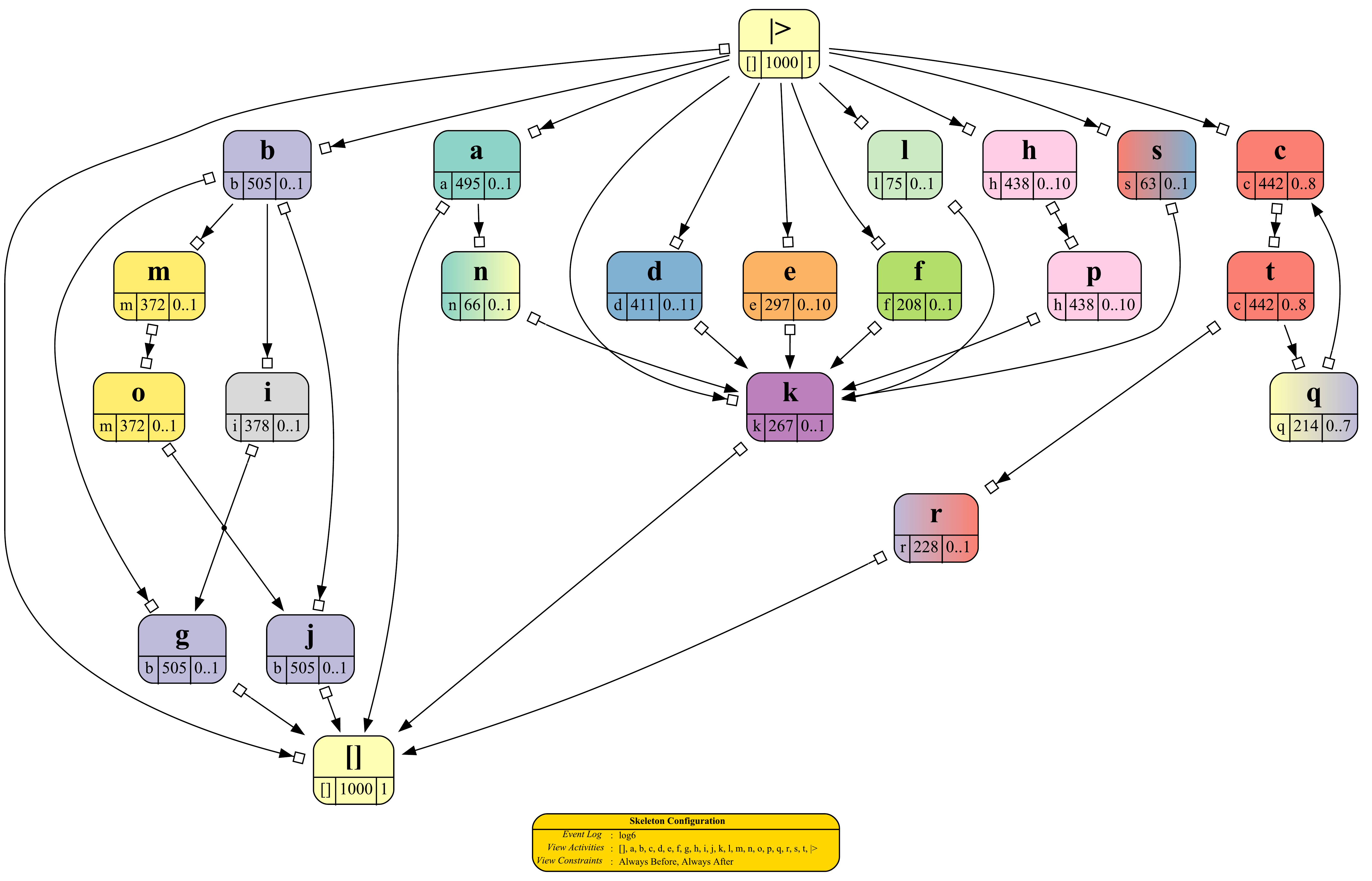}
\end{center}

\subsection{Log skeleton with hyper arcs for log $L^0_7$}
\begin{center}
\includegraphics[width=0.611\textwidth,trim={0px 80px 0px 10px},clip]{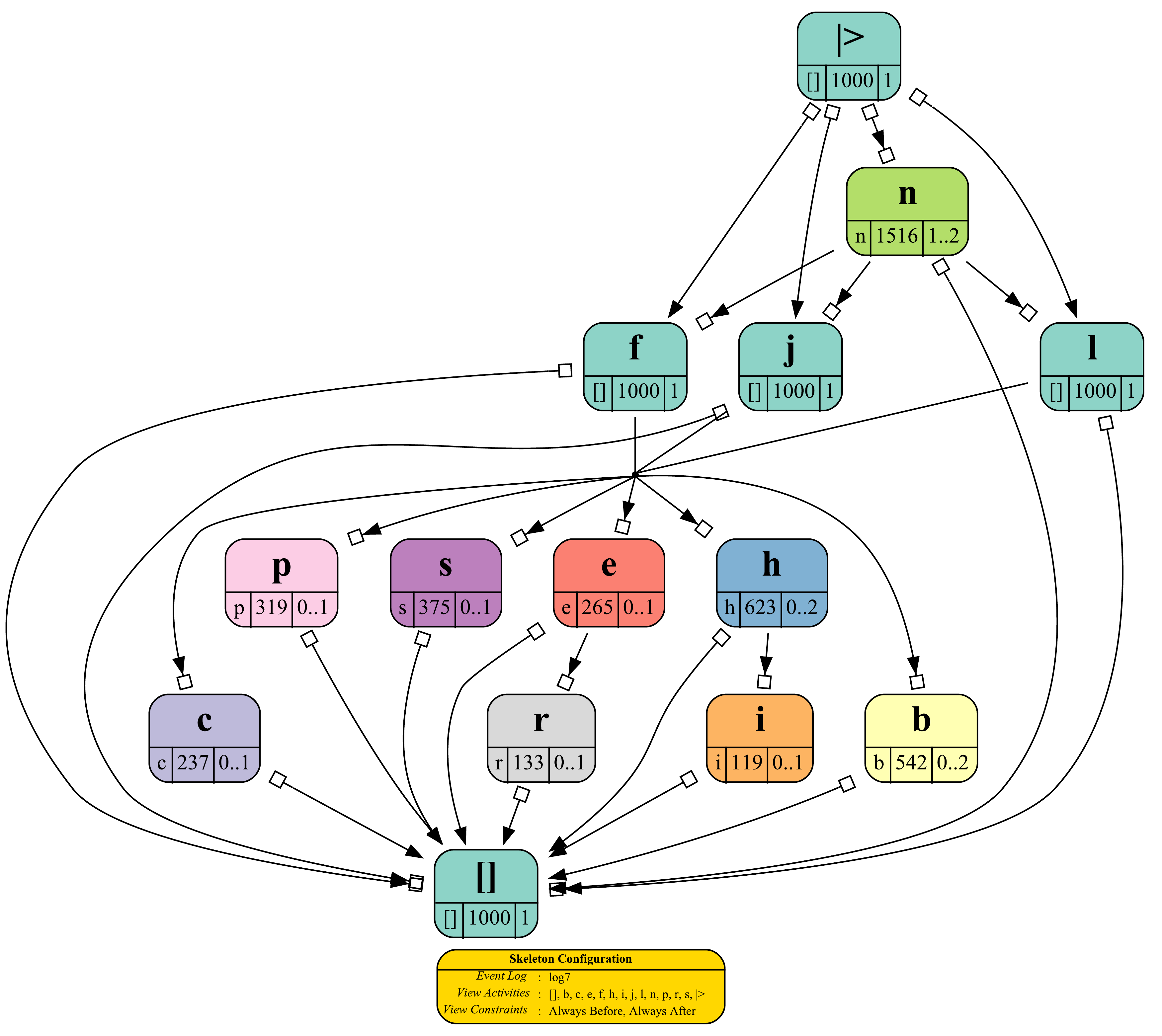}
\end{center}

\subsection{Log skeleton with hyper arcs for log $L^0_9$}
\begin{center}
\includegraphics[width=0.542\textwidth,trim={0px 80px 0px 10px},clip]{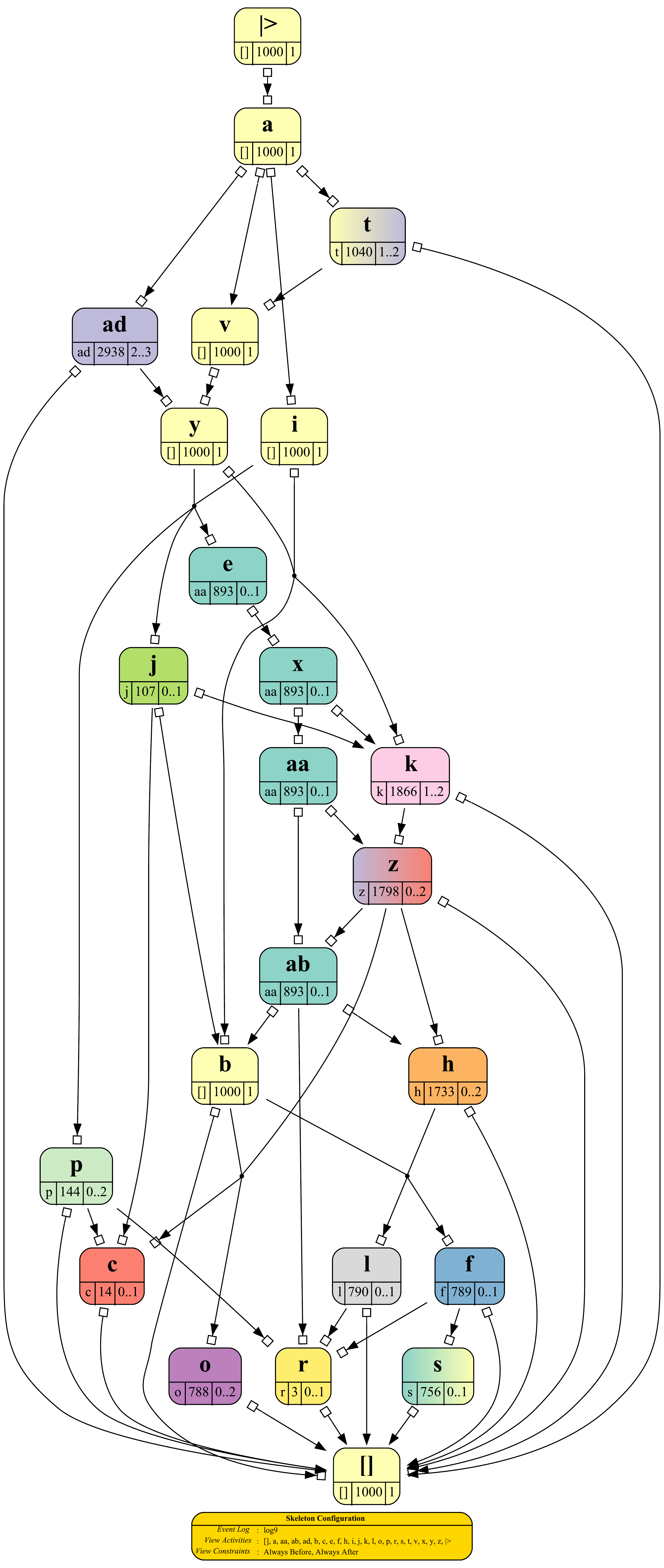}
\end{center}

\subsection{Log skeleton with hyper arcs for log $L^0_{10}$}
\begin{center}
\includegraphics[width=0.534\textwidth,trim={0px 80px 0px 10px},clip]{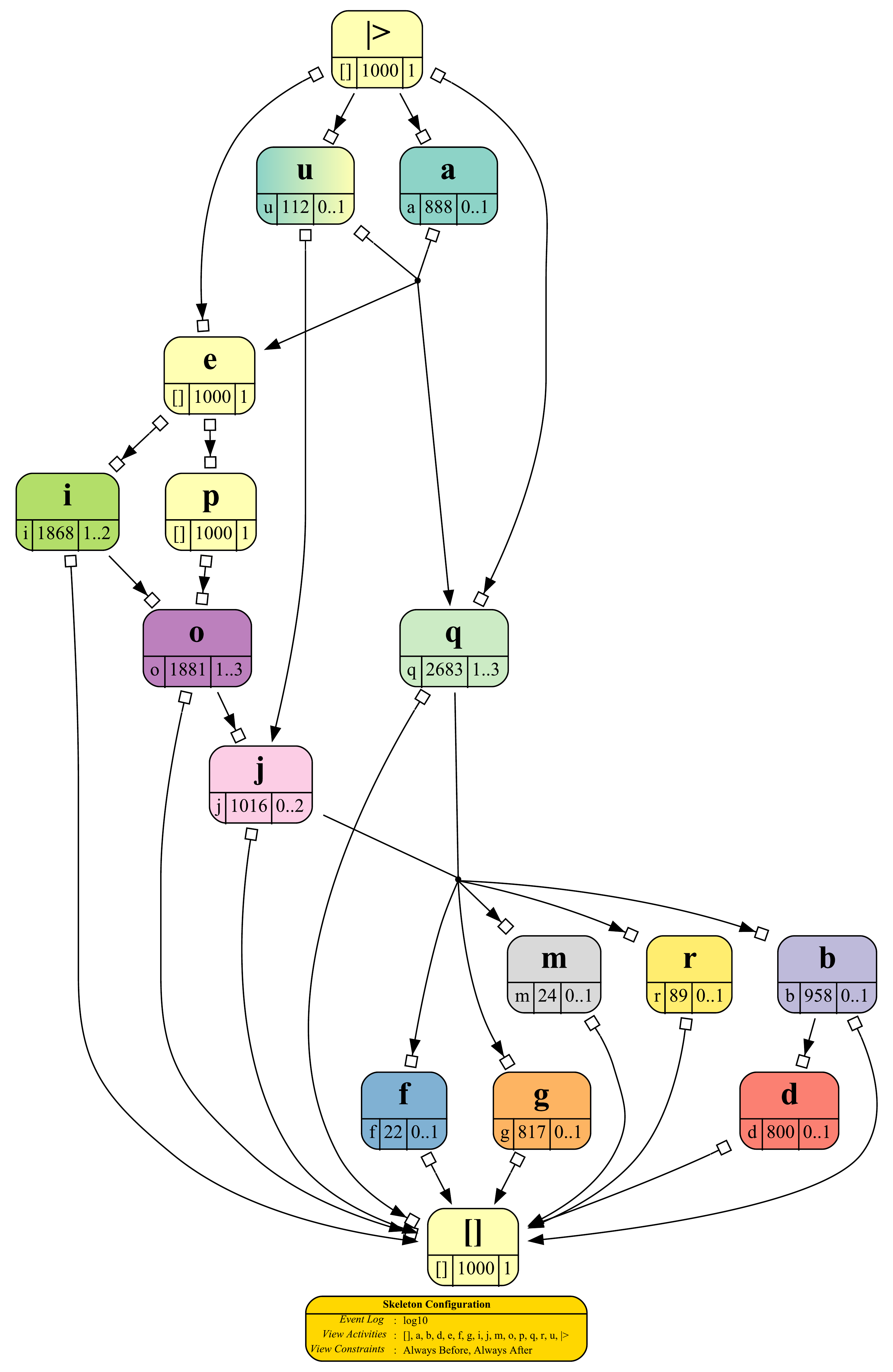}
\end{center}

\end{document}